\documentclass[10pt,twocolumn,letterpaper]{article}

\usepackage{cvpr}
\usepackage{times}
\usepackage{epsfig}
\usepackage{graphicx}
\usepackage{amsmath}
\usepackage{wrapfig}
\usepackage{amssymb}
\usepackage{url}
\usepackage{enumerate}
\usepackage{caption}
\usepackage{multirow}
\usepackage{xr}
\usepackage{adjustbox}
\usepackage{booktabs}

\usepackage{amsmath,amsfonts,bm,amsthm}

\newcommand{\norm}[1]{\left\Vert#1\right\Vert}

\newcommand{\abs}[1]{\left\vert#1\right\vert}

\newcommand{\set}[1]{\left\{#1\right\}}
\newcommand{\parr}[1]{\left (#1\right )}
\newcommand{\brac}[1]{\left [#1\right ]}

\newcommand{\Real}{\mathbb R}

\newcommand{\too}{\rightarrow}

\newcommand{\diag}{\textrm{diag}} %diagonal matrix
\newcommand{\one}{\mathbf{1}}

\newcommand{\loss}{\mathrm{loss}}%_{\text{ur}}}
\newcommand{\recloss}{\mathrm{loss}_{\text{R}}}%_{\text{ur}}}
\newcommand{\supp}{\mathrm{supp}}
\def \etal{{et al}.}
\makeatletter
\newtheorem*{rep@theorem}{\rep@title}
\newcommand{\newreptheorem}[2]{%
\newenvironment{rep#1}[1]{%
 \def\rep@title{#2 \ref{##1}}%
 \begin{rep@theorem}}%
 {\end{rep@theorem}}}
\makeatother

\newtheorem{theorem}{Theorem}
\newreptheorem{theorem}{Theorem}

\newreptheorem{lemma}{Lemma}

\newtheorem{definition}{Definition}

\usepackage{amsmath,amsfonts,bm}

\def\eqref#1{equation~\ref{#1}}
\def\1{\bm{1}}

\def\vmu{{\bm{\mu}}}
\def\vSigma{\bm{\Sigma}}
\def\vtheta{{\bm{\theta}}}
\def\veta{{\bm{\eta}}}

\def\vb{{\bm{b}}}

\def\vn{{\bm{n}}}

\def\vr{{\bm{r}}}

\def\vu{{\bm{u}}}
\def\vv{{\bm{v}}}
\def\vw{{\bm{w}}}
\def\vx{{\bm{x}}}
\def\vy{{\bm{y}}}
\def\vz{{\bm{z}}}

\def\mA{{\bm{A}}}

\def\mR{{\bm{R}}}

\def\mW{{\bm{W}}}
\def\mX{{\bm{X}}}
\def\mY{{\bm{Y}}}

\DeclareMathAlphabet{\mathsfit}{\encodingdefault}{\sfdefault}{m}{sl}
\SetMathAlphabet{\mathsfit}{bold}{\encodingdefault}{\sfdefault}{bx}{n}

\def\gN{{\mathcal{N}}}

\def\gP{{\mathcal{P}}}

\def\gS{{\mathcal{S}}}

\def\gX{{\mathcal{X}}}

\newcommand{\E}{\mathbb{E}}

\DeclareMathOperator*{\argmin}{arg\,min}

\usepackage[pagebackref=true,breaklinks=true,letterpaper=true,colorlinks,bookmarks=false]{hyperref}

\cvprfinalcopy % *** Uncomment this line for the final submission

\ifcvprfinal\fi
\begin{document}

%%%%%%%%% TITLE
\title{SAL: Sign Agnostic Learning of Shapes from Raw Data}

\author{Matan Atzmon \qquad Yaron Lipman \\
Weizmann Institute of Science \\
}


% \maketitle
%\thispagestyle{empty}

\twocolumn[{%
\renewcommand\twocolumn[1][]{#1}%
\maketitle
\begin{center}
    \centering
    \includegraphics[width=\textwidth]{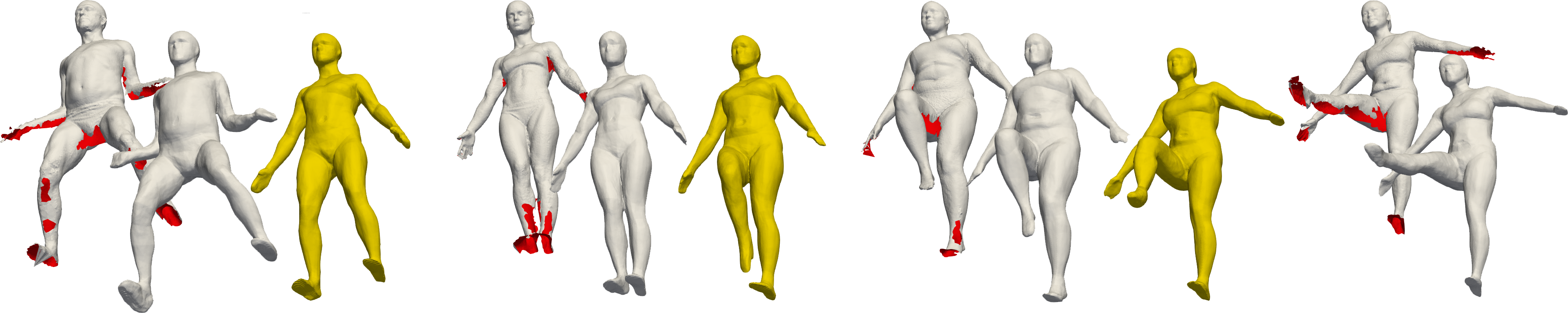}
    \captionof{figure}{We introduce SAL: Sign Agnostic Learning for learning shapes directly from raw data, such as triangle soups (left in each gray pair; back-faces are in red). Right in each gray pair - the surface reconstruction by SAL of test raw scans; in gold - SAL latent space interpolation between adjacent gray shapes. Raw scans are from the D-Faust dataset \cite{dfaust:CVPR:2017}.} \label{fig:teaser}
\end{center}%
}]

%%%%%%%%% ABSTRACT
\begin{abstract}
Recently, neural networks have been used as implicit representations for surface reconstruction, modelling, learning, and generation.  So far, training neural networks to be implicit representations of surfaces required training data sampled from a ground-truth signed implicit functions such as signed distance or occupancy functions, which are notoriously hard to compute. 

In this paper we introduce Sign Agnostic Learning (SAL), a deep learning approach for learning implicit shape representations directly from raw, unsigned geometric data, such as point clouds and triangle soups. 

We have tested SAL on the challenging problem of surface reconstruction from an un-oriented point cloud, as well as end-to-end human shape space learning directly from raw scans dataset, and achieved state of the art reconstructions compared to current approaches. We believe SAL opens the door to many geometric deep learning applications with real-world data, alleviating the usual painstaking, often manual pre-process. 

\end{abstract}

%%%%%%%%% INTRODUCTION
\section{Introduction}

Recently, deep neural networks have been used to reconstruct, learn and generate 3D surfaces. There are two main approaches: parametric \cite{groueix2018papier,ben2018multi,williams2019deep,deprelle2019learning} and implicit \cite{chen2019learning,Park_2019_CVPR,mescheder2019occupancy,atzmon2019controlling,deng2019cvxnets,genova2019learning}. In the parametric approach neural nets are used as parameterization mappings, while the implicit approach represents surfaces as zero level-sets of neural networks:
\begin{equation}\label{e:gM}
    \gS = \set{\vx\in \Real^3 \ \vert \ f(\vx;\vtheta)=0}, 
\end{equation}
where $f:\Real^{3}\times \Real^m \too \Real $ is a neural network, \eg, multilayer perceptron (MLP). The benefit in using neural networks as implicit representations to surfaces stems from their flexibility and approximation power (\eg, Theorem 1 in \cite{atzmon2019controlling}) as well as their efficient optimization and generalization properties.

So far, neural implicit surface representations were mostly learned using a regression-type loss, requiring data samples from a ground-truth implicit representation of the surface, such as a signed distance function \cite{Park_2019_CVPR} or an occupancy function \cite{chen2019learning,mescheder2019occupancy}. Unfortunately, for the common raw form of acquired 3D data $\gX\subset \Real^3$, \ie, a point cloud or a triangle soup\footnote{A triangle soup is a collection of triangles in space, not necessarily consistently oriented or a manifold.}, no such data is readily available and computing an implicit ground-truth representation for the underlying surface is a notoriously difficult task \cite{berger2017survey}. 

In this paper we advocate \emph{Sign Agnostic Learning} (SAL), defined by a family of loss functions that can be used directly with raw (unsigned) geometric data $\gX$ and produce \emph{signed} implicit representations of surfaces. An important application for SAL is in generative models such as variational auto-encoders \cite{kingma2013auto}, learning shape spaces directly from the raw 3D data. Figure \ref{fig:teaser} depicts an example where collectively learning a dataset of raw human scans using SAL overcomes many imperfections and artifacts in the data (left in every gray pair) and provides high quality surface reconstructions (right in every gray pair) and shape space (interpolations of latent representations are in gold).

We have experimented with SAL for surface reconstruction from point clouds as well as learning a human shape space from the raw scans of the D-Faust dataset \cite{dfaust:CVPR:2017}. Comparing our results to current approaches and baselines we found SAL to be the method of choice for learning shapes from raw data, and believe SAL could facilitate many computer vision and computer graphics shape learning applications, allowing the user to avoid the tedious and unsolved problem of surface reconstruction in preprocess. Our code is available at \url{https://github.com/matanatz/SAL}.

%\begin{figure}
%    \begin{tabular}{@{\hskip 0pt}c@{\hskip 0pt}c@{\hskip 0pt}c@{\hskip 0pt}}
%     \includegraphics[width=0.33\columnwidth]{figures/zero_levelset_init.pdf} & 
%     \includegraphics[width=0.33\columnwidth]{figures/zero_levelset_L0.pdf} & 
%     \includegraphics[width=0.33\columnwidth]{figures/zero_levelset_L2.pdf} \\
%    (a) & (b) & (c)
%    \end{tabular}
%    \caption{We design a family of loss functions and geometric weight initialization to reconstruct and model raw data such as point clouds and triangle soups as zero level-sets of signed distance measures. In (a) we show the initial 2D point cloud (gray) and a contour plot of an MLP using our geometric initialization. (b) shows the result of optimizing our loss with an unsigned $L^0$ distance to the point cloud; and (c) with an unsigned $L^2$ distances. We depict the zero level-sets in bold.}
%    \label{fig:2d}
%%    \vspace{-10pt}
%\end{figure}

\section{Previous work}

\subsection{Surface learning with neural networks}

%Previous work aiming at surface approximation have mostly solved the manifold approximation problem by regressing either the \emph{signed distance function} (SDF) \cite{Park_2019_CVPR} or the shape inclusion function (IF) \cite{chen2019learning} \yl{add all, also our method}. In all cases the SDF and/or IF values at regression points $\gY\subset \Real^d$ are provided as train data for learning $f$ using a standard regression loss. The challenge is that SDF/IF values are often hard to compute since they require global manifold reconstruction or at-least consistent oriented normals which are a challenge due to topological or other noise, sampling issues, and complicated manifold geometry. 

\paragraph{Neural parameteric surfaces.}
One approach to represent surfaces using neural networks is parametric, namely, as parameterization charts $f:\Real^2\too \Real^3$. Groueix \etal ~\cite{groueix2018papier} suggest to represent a surface using a collection of such parameterization charts (\ie, atlas); Williams \etal ~\cite{williams2019deep} optimize an atlas with proper transition functions between charts and concentrate on reconstructions of individual surfaces. Sinha \etal~\cite{sinha2016deep,sinha2017surfnet} use geometry images as global parameterizations, while  \cite{maron2017convolutional} use conformal global parameterizations to reduce the number of degrees of freedom of the map. Parametric representation are explicit but require handling of coverage, overlap and distortion of charts.

\paragraph{Neural implicit surfaces.}
Another approach to represent surfaces using neural networks, which is also the approach taken in this paper, is using an implicit representation, namely $f:\Real^3\too\Real$ and the surface is defined as its zero level-set, \eqref{e:gM}. Some works encode $f$ on a volumetric grid such as voxel grid \cite{wu2016learning} or an octree \cite{tatarchenko2017octree}. More flexibility and potentially more efficient use of the degrees of freedom of the model are achieved when the implicit function $f$ is represented as a neural network \cite{chen2019learning,Park_2019_CVPR,mescheder2019occupancy,atzmon2019controlling,genova2019learning}.  In these works the implicit is trained using a regression loss of the signed distance function \cite{Park_2019_CVPR}, an occupancy function \cite{chen2019learning,mescheder2019occupancy} or via particle methods to directly control the neural level-sets \cite{atzmon2019controlling}. Excluding the latter that requires sampling the zero level-set, all regression-based methods require ground-truth inside/outside information to train the implicit $f$. In this paper we present a sign agnostic training method, namely training method that can work directly with the raw (unsigned) data. 

%The main benefit in this representation is that it requires a single scalar function $f$ to represent an arbitrary surface, in contrast to an atlas that requires covering and transitions. The main drawback is that it does not give an explicit parametric representation to the surface and sampling the surface requires finding zeros of $f$. 

\pagebreak
\paragraph{Shape representation learning.} Learning collections of shapes is done using Generative Adversarial Networks (GANs) \cite{goodfellow2014generative}, auto-encoders and variational auto-encoders  \cite{kingma2013auto}, and auto-decoders \cite{bojanowski2017optimizing}. Wu \etal~\cite{wu2016learning} use GAN on a voxel grid encoding of the shape, while Ben-Hamu \etal~\cite{ben2018multi} apply GAN on a collection of conformal charts. Dai \etal~\cite{dai2017shape} use encoder-decoder architecture to learn a signed distance function to a complete shape from a partial input on a volumetric grid. Stutz \etal~\cite{stutz2018learning} use variational auto-encoder to learn an implicit surface representations of cars using a volumetric grid. Baqautdinov \etal~\cite{bagautdinov2018modeling} use variational auto-encoder with a constant mesh to learn parametrizations of faces shape space. Litany \etal~\cite{litany2018deformable} use variational auto-encoder to learn body shape embeddings of a template mesh. Park \etal~\cite{Park_2019_CVPR} use auto-decoder to learn implicit neural representations of shapes, namely directly learns a latent vector for every shape in the dataset. In our work we also make use of a variational auto-encoder but differently from previous work, learning is done directly from raw 3D data. 

\subsection{Surface reconstruction.} 

\paragraph{Signed surface reconstruction.}
Many surface reconstruction methods require normal or inside/outside information.  Carr \etal~\cite{carr2001reconstruction} were among the first to suggest using a parametric model to reconstruct a surface by computing its implicit representation; they use radial basis functions (RBFs) and regress at inside and outside points computed using oriented normal information. Kazhdan \etal~\cite{kazhdan2006poisson,kazhdan2013screened} solve a Poisson equation on a volumetric discretization to extend points and normals information to an occupancy indicator function. Walder \etal~\cite{walder2007implicit} use radial basis functions and solve a variational hermite problem (\ie, fitting gradients of the implicit to the normal data) to avoid trivial solution.  In general our method works with a non-linear parameteric model (MLP) and therefore does not require a-priori space discretization nor works with a fixed linear basis such as RBFs.

\paragraph{Unsigned surface reconstruction.}
More related to this paper are surface reconstruction methods that work with unsigned data such as point clouds and triangle soups. 
Zhao \etal~\cite{zhao2001fast} use the level-set method to fit an implicit surface to an unoriented point cloud by minimizing a loss penalizing distance of the surface to the point cloud achieving a sort of minimal area surface interpolating the points. Walder \etal~\cite{walder2005implicit} formulates a variational problem fitting an implicit RBF to an unoriented point cloud data while minimizing a regularization term and maximizing the norm of the gradients; solving the variational problem is equivalent to an eigenvector problem.  
Mullen \etal~\cite{mullen2010signing} suggests to sign an unsigned distance function to a point cloud by a multi-stage algorithm first dividing the problem to near and far field sign estimation, and propagating far field estimation closer to the zero level-set; then optimize a convex energy fitting a smooth sign function to the estimated sign function. Takayama \etal~\cite{takayama2014consistently} suggested to orient triangle soups by minimizing the Dirichlet energy of the generalized winding number noting that correct orientation yields piecewise constant winding number.  Xu \etal~\cite{xu2014signed} suggested to compute robust signed distance function to triangle soups by using an offset surface defined by the unsigned distance function.  Zhiyang \etal~\cite{Zhiyang2019} fit an RBF implicit by optimizing a non-convex variational problem minimizing smoothness term, interpolation term and unit gradient at data points term. 
All these methods use some linear function space; when the function space is global, \eg when using RBFs, model fitting and evaluation are costly and limit the size of point clouds that can be handled efficiently, while local support basis functions usually suffer from inferior smoothness properties \cite{wendland2004scattered}. In contrast we use a non-linear function basis (MLP) and advocate a novel and simple sign agnostic loss to optimize it. Evaluating the non-linear neural network model is efficient and scalable and the training process can be performed on a large number of points, \eg, with stochastic optimization techniques.

\begin{figure}
\begin{tabular}{@{\hskip 0.03\columnwidth}c@{\hskip 0.03\columnwidth}c@{\hskip 0.03\columnwidth}}
\includegraphics[width=0.46\columnwidth]{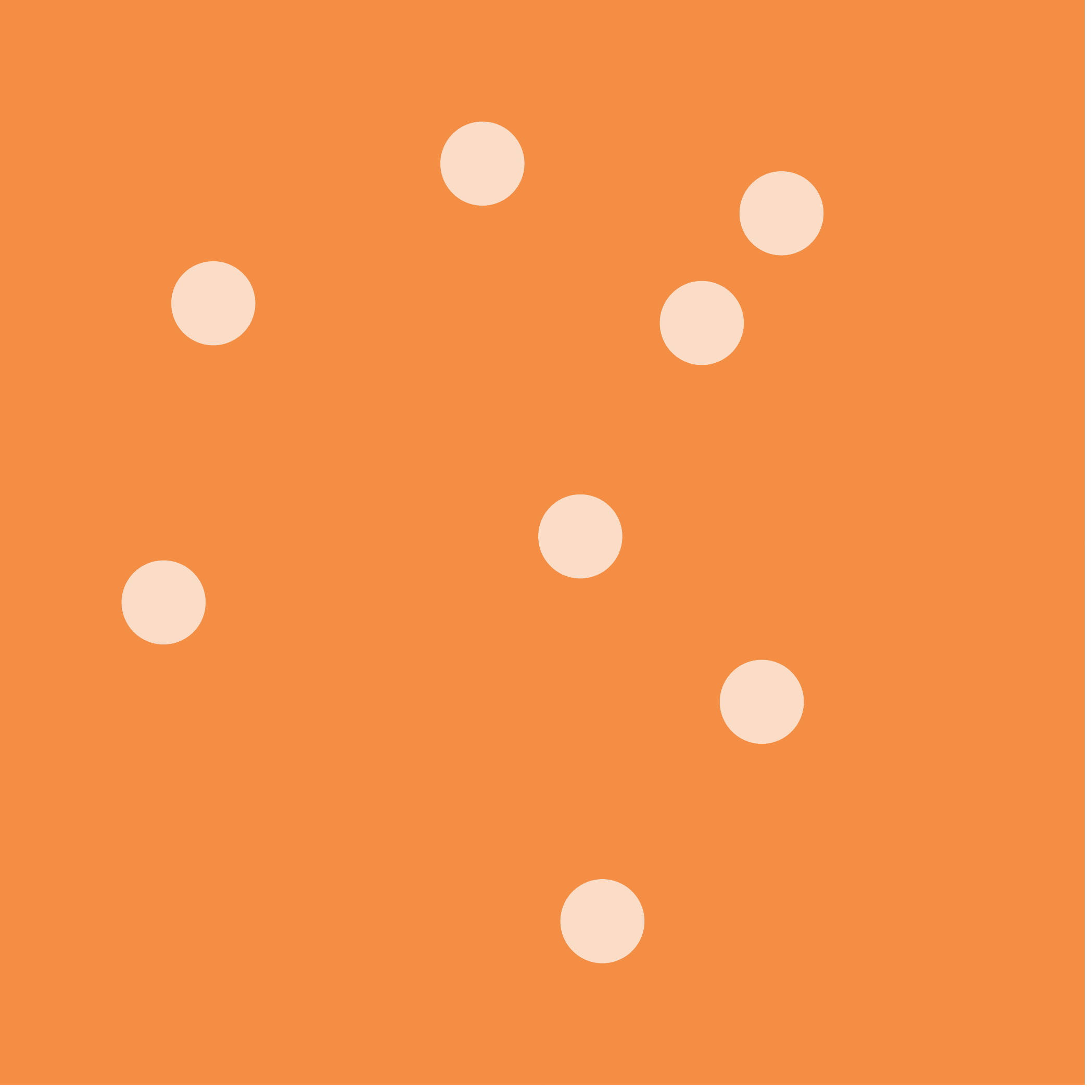} &
\includegraphics[width=0.46\columnwidth]{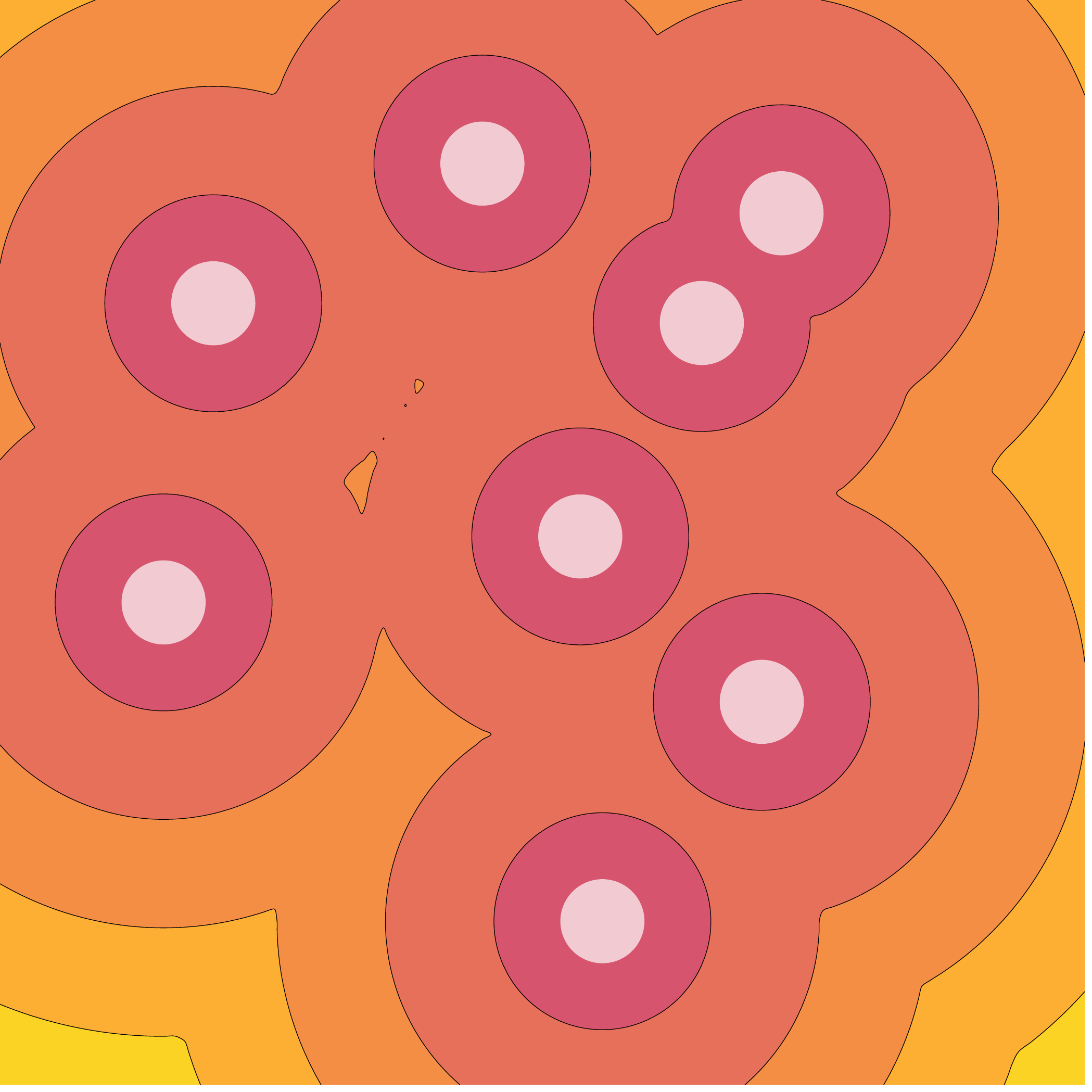} \\ 
(a) & (b) \\
\includegraphics[width=0.46\columnwidth]{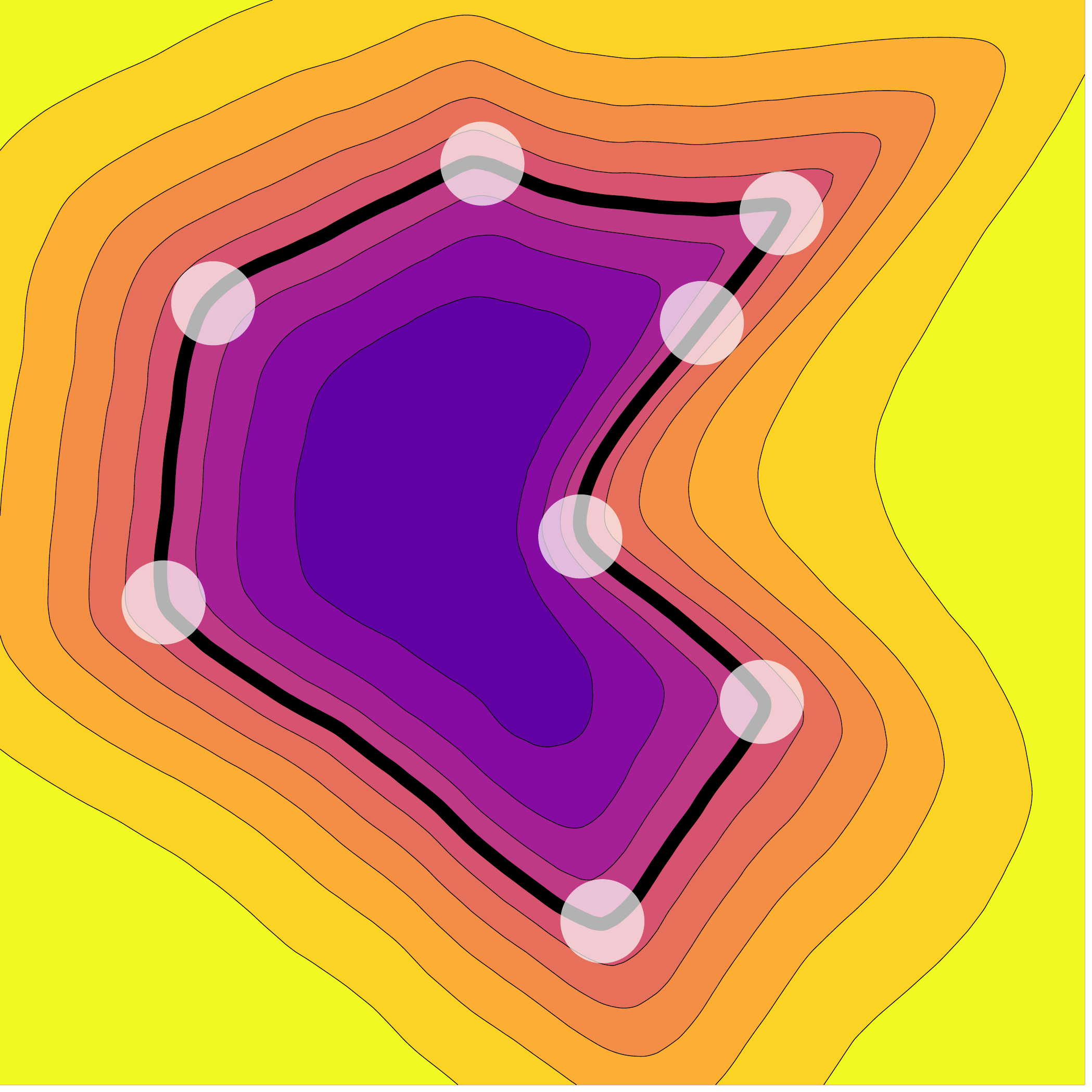} &
\includegraphics[width=0.46\columnwidth]{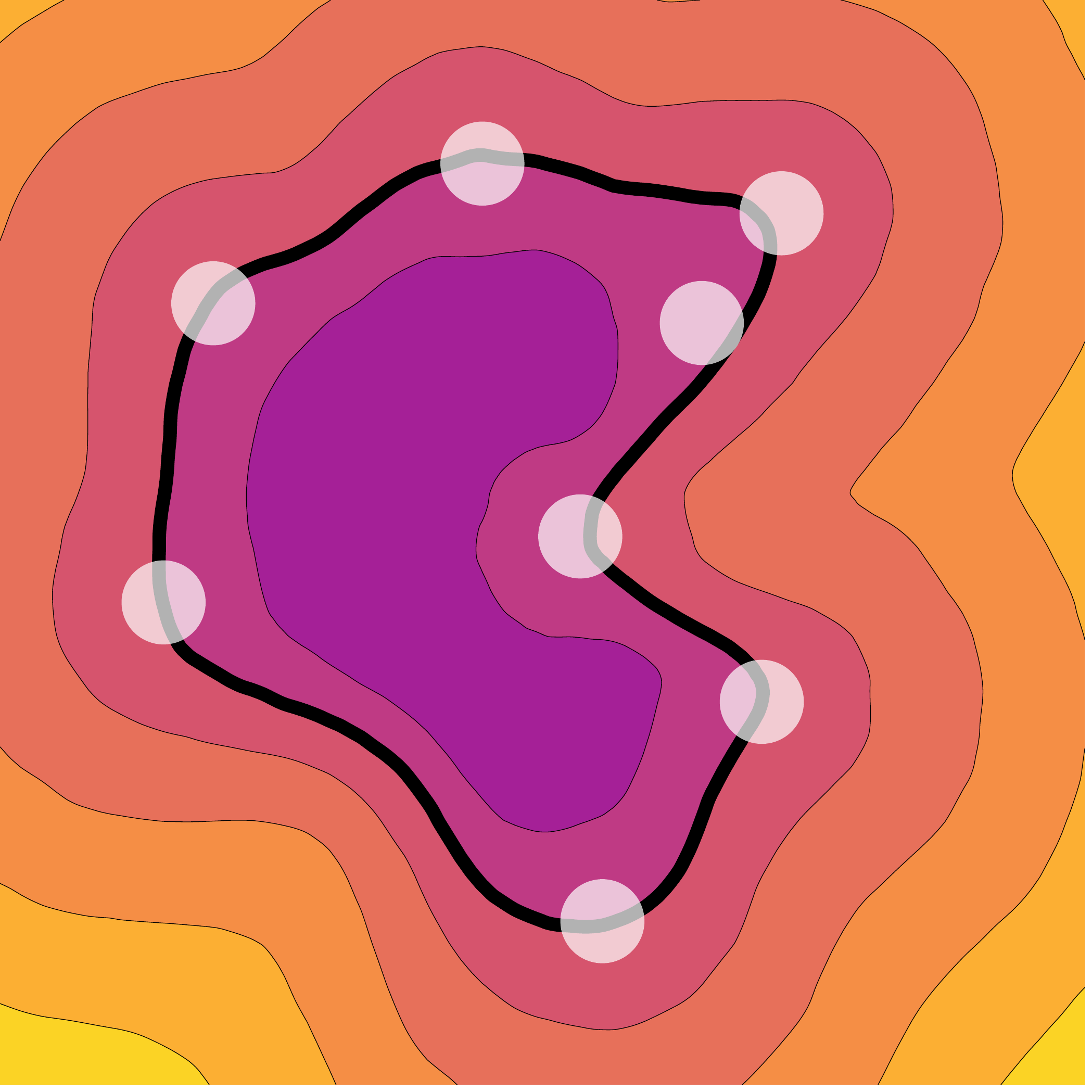} \\
(c) & (d) \\
\end{tabular}
\caption{Experiment with sign agnostic learning in 2D:  (a) and (b) show the unsigned $L^0$ and $L^2$ (resp.)~distances to a 2D point cloud (in gray); (c) and (d) visualize the different level-sets of the neural networks optimized with the respective sign agnostic losses. Note how the zero level-sets (in bold) gracefully connect the points to complete the shape. }
\label{fig:2d}
%    \vspace{-10pt}
\end{figure}

\section{Sign agnostic learning}
%\yl{elaborate explanation why we find signed solution, what initialisation got to do with it. suggested: 1) the SAL energy has both the signed and unsigned solutions as local minima. 
%2) to get to the signed local minima we empirically noticed that one needs to initialize the network from some rough signed distance function. This is why we calculated weight initialisation to get approximate spherical signed distance distance. }

Given a raw input geometric data,  $\gX \subset \Real^3$, \eg, a point cloud or a triangle soup, we are looking to optimize the weights $\vtheta\in\Real^m$ of a network $f(\vx;\vtheta)$, where $f:\Real^3\times\Real^m\too\Real$, so that its zero level-set, \eqref{e:gM}, is a surface approximating $\gX$. 

We introduce the \emph{Sign Agnostic Learning} (SAL) defined by a loss of the form
\begin{equation}\label{e:loss}
    \loss (\vtheta)= \E_{\vx\sim D_\gX} \  \tau \Big ( f(\vx;\vtheta) , h_\gX(\vx) \Big ),
\end{equation}
where $D_\gX$ is a probability distribution defined by the input data $\gX$; $h_\gX(\vx)$ is some \emph{unsigned} distance measure to $\gX$; and $\tau:\Real\times \Real_+\too \Real$ is a differentiable \emph{unsigned similarity function} defined by the following properties: 
\begin{enumerate}[(i)]
    \item \emph{Sign agnostic:} $\tau(-a,b)=\tau(a,b)$,  $\forall a\in \Real$, $b\in \Real_+$.
    % \mathrm{image}(h_\gX)$. 
    \label{item:sym}
  % \item \emph{Monotonic:} $\sign(\frac{\partial \tau}{ \partial a}(a,b))\hspace{-3pt}=\hspace{-2pt}\sign(a-b)$, $\forall a,b\in \Real_+$.
   
    \item \emph{Monotonic:} $\frac{\partial \tau}{ \partial a}(a,b) = \rho(a-b)$, $\forall a,b\in\Real_+$,
    
    \label{item:mono}
\end{enumerate}
where $\rho:\Real\too\Real$ is a monotonically increasing function with $\rho(0)=0$.
An example of an unsigned similarity is $\tau(a,b)=\abs{\abs{a}-b}$. 

To understand the idea behind the definition of the SAL loss, consider first a standard regression loss using $\tau(a,b)=\abs{a-b}$ in \eqref{e:loss}. This would encourage $f$ to resemble the unsigned distance $h_\gX$ as much as possible.  On the other hand, using the unsigned similarity $\tau$ in \eqref{e:loss} introduces a new local minimum of $\loss$ where $f$ is a \emph{signed} function such that $\abs{f}$ approximates $h_\gX$. To get this desirable local minimum we later design a network weights' initialization $\vtheta^0$ that favors the signed local minima.

\begin{wrapfigure}[6]{r}{0.25\columnwidth}
\vspace{-12pt}
\hspace{-15pt}
\includegraphics[width=0.28\columnwidth]{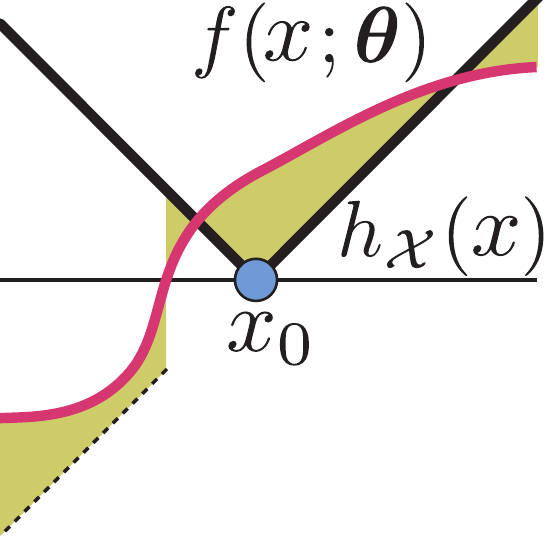}
\vspace{0pt}
%\caption{XXX}
\label{fig:wl_example}
\end{wrapfigure} 
As an illustrative example, the inset depicts the one dimensional case ($d=1$) where $\gX=\set{x_0}$, $h_\gX(x)=|x-x_0|$, and $\tau(a,b)=\abs{\abs{a}-b}$, which satisfies properties (\ref{item:sym}) and (\ref{item:mono}), as discussed below; the loss therefore strives to minimize the area of the yellow set.  
When initializing the network parameters $\vtheta=\vtheta^0$ properly, the minimizer $\vtheta^*$ of $\loss$ defines an implicit $f(x;\vtheta^*)$  that realizes a \emph{signed} version of $h_\gX$; in this case $f(x;\vtheta^*)=x-x_0$. In the three dimensional case the zero level-set $\gS$ of $f(\vx;\vtheta^*)$ will represent a surface approximating $\gX$.  

To theoretically motivate the loss family in \eqref{e:loss} we will prove that it possess a \emph{plane reproduction} property. That is, if the data $\gX$ is contained in a plane, there is a critical weight $\vtheta^*$ reconstructing this plane as the zero level-set of $f(\vx;\vtheta^*)$. Plane reproduction is important for surface approximation since surfaces, by definition, have an approximate tangent plane almost everywhere \cite{do2016differential}. 

We will explore instantiations of SAL based on different choices of unsigned distance functions $h_\gX$, as follows.

\paragraph{Unsigned distance functions.} We consider two $p$-distance functions: For $p=2$ we have the standard $L^2$ (Euclidean) distance 
\begin{equation}\label{e:L2_dist}
 h_2(\vz) = \min_{\vx\in\gX}\norm{\vz-\vx}_2,
\end{equation}
and for $p=0$ the $L^0$ distance
\begin{equation}\label{e:L0_dist}
  h_0(\vz)=\begin{cases} 0 & \vz\in\gX \\ 1 & \vz\notin \gX \end{cases}.  
\end{equation}

\paragraph{Unsigned similarity function.}
Although many choices exist for the unsigned similarity function, in this paper we take
\begin{equation}\label{e:tau}
    \tau_\ell(a,b) = \abs{\abs{a}-b}^\ell,
\end{equation}
where $\ell\geq 1$. The function $\tau_\ell$ is indeed an unsigned similarity: it satisfies (\ref{item:sym}) due to the symmetry of $\abs{\cdot}$; and since $\frac{\partial \tau}{\partial a} = \ell\abs{|a|-b}^{\ell-1}\mathrm{sign}(a-b\,\mathrm{sign}(a))$ it satisfies (\ref{item:mono}) as-well.

\paragraph{Distribution $D_\gX$.}
The choice of $D_\gX$ is depending on the particular choice of $h_\gX$. For $L^2$ distance, it is enough to make the simple choice of splatting an isotropic Gaussian, $\gN(\vx,\sigma^2 I)$, at every point (uniformly randomized) $\vx\in\gX$; we denote this probability $\gN_\sigma(\gX)$; note that $\sigma$ can be taken to be a function of $\vx\in\gX$ to reflect local density in $\gX$. In this case, the loss takes the form
\begin{equation}\label{e:loss_L2}
\loss(\vtheta) =  \E_{\vz\sim \gN_\sigma(\gX)}\big| |f(\vz;\vtheta)| - h_2(\vz) \big|^\ell. 
% \loss(\vtheta) = \frac{1}{n} \sum_{i=1}^n \big| |f(\vy_i;\vtheta)| - h_2(\vy_i) \big|^\ell,
%\loss(\vtheta) = \frac{1}{n} \sum_{i=1}^n \big| |f(\vy_i;\vtheta)| - h_2(\vy_i) \big|^\ell,
\end{equation}
%where $\gN(\gX,\sigma I)$ denotes the distribution of convolving a Gaussian  $\vy_i \sim \gN(\vx_i,\sigma^2 I)$. 

For the $L^0$ distance however, $h_\gX(\vx)\ne 1$ only for $\vx\in\gX$ and therefore a non-continuous density should be used; we opt for $\gN(\vx,\sigma^2 I) + \delta_\vx$, where $\delta_\vx$ is the delta distribution measure concentrated at $\vx$. The loss takes the form
\begin{equation}\label{e:loss_L0}
\loss(\vtheta) = \E_{\vz\sim \gN_\sigma(\gX)} \big| |f(\vz;\vtheta)| - 1 \big|^\ell + \E_{\vx \sim \gX}\big| f(\vx;\vtheta) \big|^\ell. 
%\loss(\vtheta) = \frac{1}{n} \sum_{i=1}^n \big| |f(\vy_i;\vtheta)| - 1 \big|^\ell + \frac{1}{n}\sum_{i=1}^n \big| f(\vx_i;\vtheta) \big|^\ell,
\end{equation}
%where $\vy_i \sim \gN(\vx_i,\sigma^2 I)$. 
Remarkably, the latter loss requires only randomizing points $\vz$ near the data samples without any further computations involving $\gX$. This allows processing of large and/or complex geometric data.

\paragraph{Neural architecture.}
Although SAL can work with different parametric models, in this paper we consider a multilayer perceptron (MLP) defined by \begin{equation}\label{e:f}
f(\vx;\vtheta)= \varphi \big( \vw^T f_{\ell}\circ f_{\ell-1} \circ \cdots \circ f_1 (\vx) + b \big ),\end{equation}
and
\begin{equation}\label{e:f_params}
 f_i(\vy) = \nu(\mW_i \vy +\vb_i), \mW \in \Real^{d_i^{out}\times d_i^{in}},  \vb_i\in\Real^{d_i^{out}},
\end{equation}
where $\nu(a)=(a)_+$ is the ReLU activation, and $\vtheta = (\vw,b,\mW_\ell,\vb_\ell,\ldots,\mW_1,\vb_1)$; $\varphi$ is a strong non-linearity, as defined next:
\begin{definition}
The function $\varphi:\Real\too\Real$ is called a \emph{strong non-linearity} if it is differentiable (almost everywhere), anti-symmetric, $\varphi(-a)=-\varphi(a)$, and there exists $\beta\in\Real_+$ so that $\beta^{-1}\geq \varphi'(a) \geq \beta > 0$, for all $a\in\Real$ where it is defined.
\end{definition}
In this paper we use $\varphi(a)=a$ or $\varphi(a)=\tanh(a)+\gamma a$, where $\gamma\geq 0$ is a parameter. Furthermore, similarly to previous work \cite{Park_2019_CVPR,chen2019learning} we have incorporated a skip connection layer $s$, concatenating the input $\vx$ to the middle hidden layer, that is  $s(\vy)=(\vy,\vx)$, where here $\vy$ is a hidden variable in $f$.

\paragraph{2D example.}
%\begin{wrapfigure}[5]{r}{0.2\columnwidth}
%\vspace{-10pt}
%\hspace{-15pt}
%\includegraphics[width=0.21\columnwidth]{figures/zero_levelset_init.pdf}
%\vspace{0pt}
%%\caption{XXX}
%\label{fig:2d_init}
%\end{wrapfigure} 
The two examples in Figure \ref{fig:2d} show case the SAL for a 2D point cloud, $\gX=\set{\vx_i}_{i=1}^8\subset \Real^2$, (shown in gray) as input.  These examples were computed by optimizing \eqref{e:loss_L2} (right column) and \eqref{e:loss_L0} (left column) with $\ell=1$ using the $L^2$ and $L^0$ distances (resp.). 
The architecture used is an 8-layer MLP; all hidden layers are 100 neurons wide, with a skip connection to the middle layer. \\
%
%The initialization we used, $\vtheta=\vtheta^0$, is explained \yl{in the next section.}

Notice that both $h_\gX(\vx)$ and its signed version are local minima of the loss in \eqref{e:loss}. These local minima are stable in the sense that there is an energy barrier when moving from one to the other. For example, to get to a solution as in Figure \ref{fig:2d}(b) from the solution in Figure \ref{fig:2d}(d) one needs to flip the sign in the interior or exterior of the region defined by the black line. Changing the sign continuously will result in a considerable increase to the SAL loss value.\\

We elaborate on our initialization method, $\vtheta=\vtheta^0$, that in practice favors the signed version of $h_\gX$ in the next section.

% $\vtheta=\vtheta^0$ is designed to make $f$ approximate the signed distance to a sphere (in this case, a circle), as shown in the inset figure. 

\begin{figure}[t]
    \begin{tabular}{@{\hskip 0pt}c@{\hskip 0pt}c@{\hskip 0pt}c@{\hskip 0pt}}
     \includegraphics[width=0.33\columnwidth]{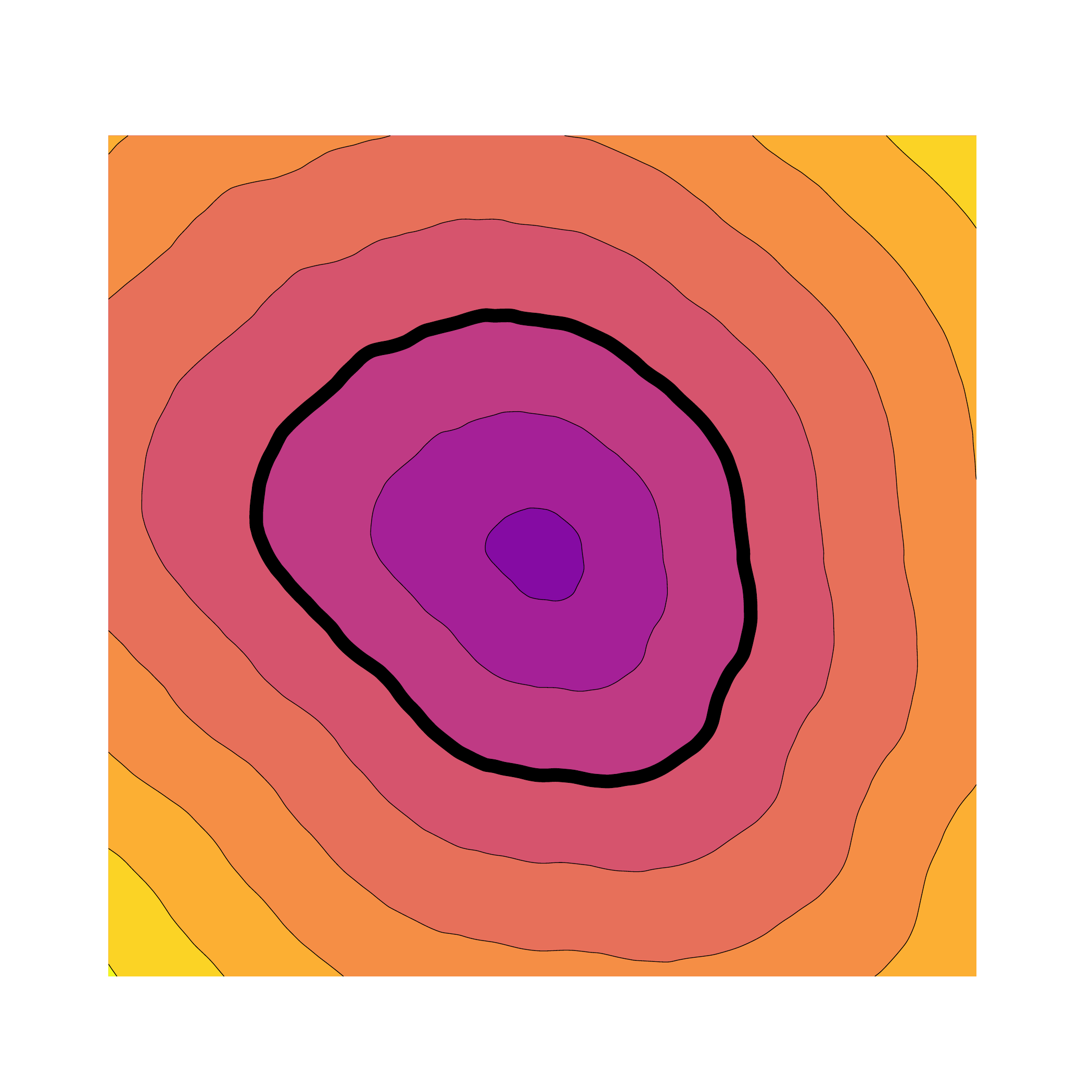} & 
     \includegraphics[width=0.33\columnwidth]{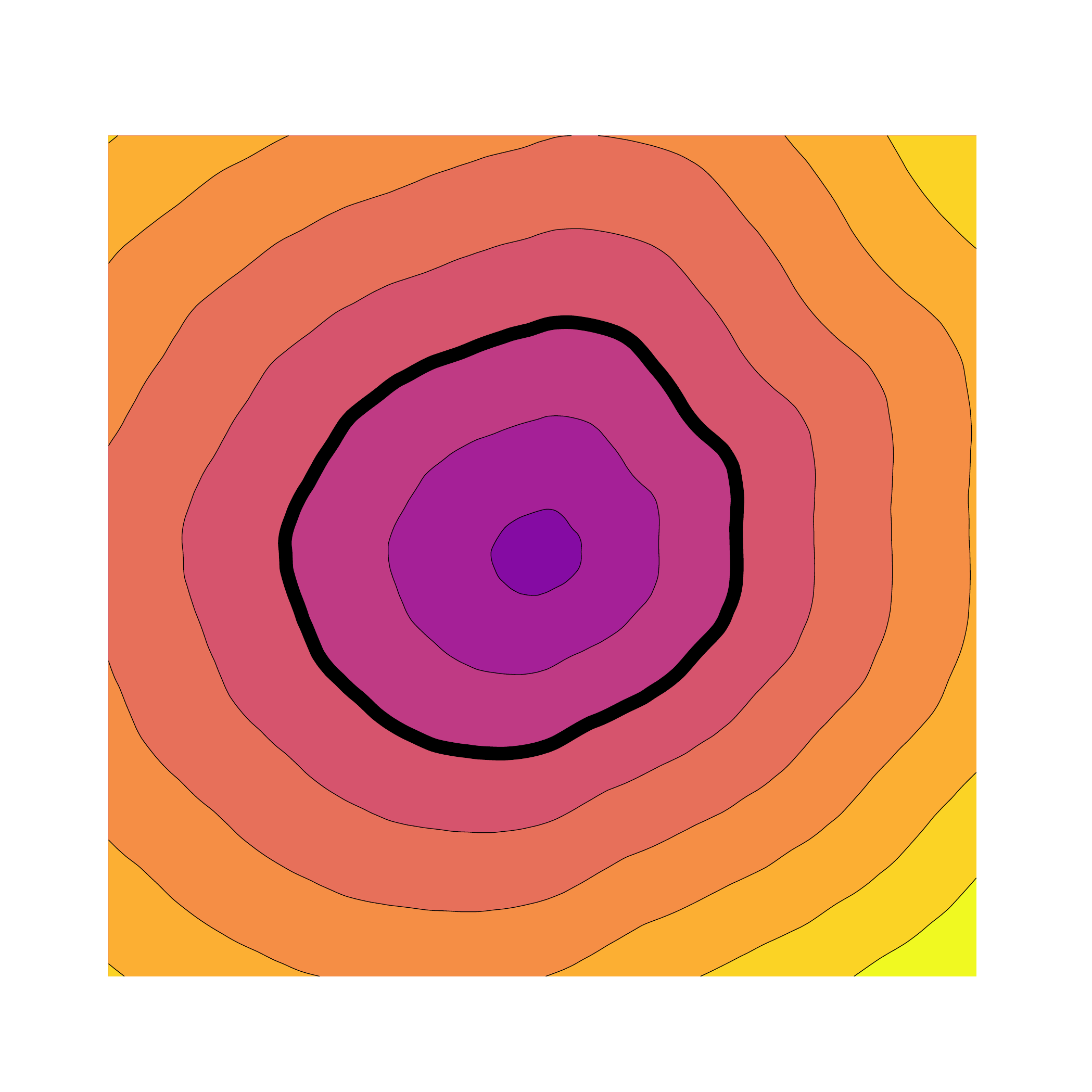} & 
     \includegraphics[width=0.33\columnwidth]{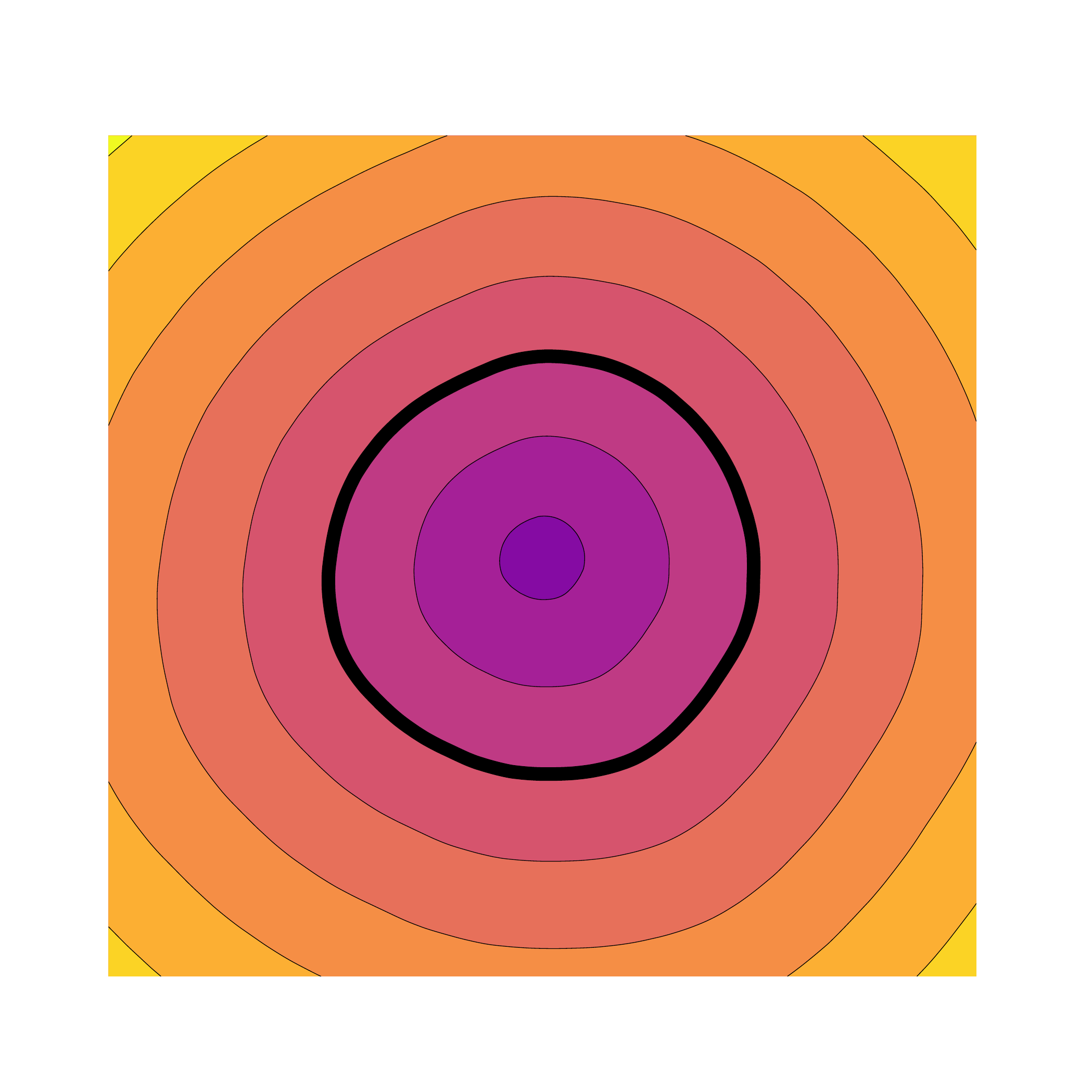} \\
    (a) & (b) & (c) 
    \end{tabular}
    \caption{Geometric initialization of neural networks: An MLP with our weight initialization (see Theorem \ref{cor:geometric_init})  is approximating the signed distance function to an $r$-radius sphere, $f(\vx;\vtheta^0)\approx \varphi(\norm{\vx}-r)$, where the approximation improves with the width of the hidden layers: (a) depicts an MLP with $100$-neuron hidden layers; (b) with $200$; and (c) with $2000$.   }
    \label{fig:2d_init}
%    \vspace{-10pt}
\end{figure}

\section{Geometric network initialization}\label{s:init}
A key aspect of our method is a proper, geometrically motivated initialization of the network's parameters. For MLPs, equations \ref{e:f}-\ref{e:f_params}, we develop an initialization of its parameters, $\vtheta=\vtheta^0$, so that $f(\vx;\vtheta^0)\approx \varphi(\norm{\vx}-r)$, where $\norm{\vx}-r$ is the signed distance function to an $r$-radius sphere. The following theorem specify how to pick $\vtheta^0$ to achieve this: 
\begin{theorem}\label{cor:geometric_init}
Let $f$ be an MLP (see equations \ref{e:f}-\ref{e:f_params}). Set, for $1\leq i \leq \ell$, $\vb_i=0$ and $\mW_i$ i.i.d.~from a normal distribution $\gN(0,\frac{\sqrt{2}}{\sqrt{d_i^{out}}})$; further set $\vw={\frac{\sqrt{\pi}}{\sqrt{d_\ell^{out}}}}\one$, $c=-r$. Then, $f(\vx)\approx \varphi(\norm{x}-r)$.
\end{theorem}
Figure \ref{fig:2d_init} depicts level-sets (zero level-sets in bold) using the initialization of Theorem \ref{cor:geometric_init}  with the same 8-layer MLP (using $\varphi(a)=a$) and increasing width of 100, 200, and 2000 neurons in the hidden layers. Note how the approximation $f(\vx;\vtheta^0)\approx \norm{\vx}-r$ improves as the layers' width increase, while the sphere-like (in this case circle-like) zero level-set remains topologically correct at all approximation levels. 

The proof to Theorem \ref{cor:geometric_init}  is provided in the supplementary material; it is a corollary of the following theorem, showing how to chose the initial weights for a single hidden layer network: 
\begin{theorem}\label{thm:3_layers_sphere}
Let $f:\Real^{d}\too \Real$ be an MLP with ReLU activation, $\nu$, and a single hidden layer. That is, $f(\vx) = \vw^T\nu(\mW \vx+ \vb)+c$, where $\mW\in\Real^{d^{out}\times d}$, $\vb\in\Real^{d^{out}}$, $\vw\in\Real^{d^{out}}$, $c\in\Real$ are the learnable parameters. If $\vb=0$, $\vw=\frac{\sqrt{2\pi}}{\sigma d^{out}}\one$, $c=-r$, $r>0$, and all entries of $\mW$ are i.i.d.~normal $\gN(0,\sigma^2)$ then $f(\vx)\approx \norm{\vx}-r$. That is, $f$ is approximately the signed distance function to a $d-1$ sphere of radius $r$ in $\Real^{d}$, centered at the origin.  
\end{theorem}

\section{Properties}\label{s:properties}
\subsection{Plane reproduction}

Plane reproduction is a key property to surface approximation methods since, in essence, surfaces are locally planar, \ie, have an approximating tangent plane almost everywhere \cite{do2016differential}. In this section we provide a theoretical justification to SAL by proving 
a plane reproduction property.  We first show this property for a linear model (\ie, a single layer MLP) and then show how this implies local plane reproduction for general MLPs. 

The setup is as follows: Assume the input data $\gX\subset \Real^d$ lies on a hyperplane $\gX\subset \gP$, where $\gP=\set{\vx\in\Real^d \ \vert \ \vn^T \vx + c  = 0}$, $\vn\in\Real^d$, $\norm{\vn}=1$, is the normal to the plane, and consider a linear model $f(\vx;\vw,b)=\varphi(\vw^T\vx+b)$. Furthermore, we make the assumption that the distribution $D_\gX$ and the distance $h_\gX$ are invariant to rigid transformations, which is common and holds in all cases considered in this paper.  
We prove existence of critical weights $(\vw^*,b^*)$ of the loss in \eqref{e:loss}, and for which the zero level-set of $f$, $f(\vx;\vw^*,b^*)=0$, reproduces $\gP$: %\yl{add proof idea?}

\begin{theorem}\label{thm:planar_reproduction}
Consider a linear model  $f(\vx;\vtheta)=\varphi(\vw^T\vx+b)$, $\vtheta=(\vw,b)$, with a strong non-linearity $\varphi:\Real\too\Real$. Assume the data $\gX$ lies on a plane $\gP=\set{\vx\vert \vn^T\vx+c=0}$, \ie, $\gX\subset \gP$. Then, there exists $\alpha\in \Real_+$ so that $(\vw^*,b^*)=(\alpha\vn, \alpha c)$ is a critical point of the loss in \eqref{e:loss}. 
\end{theorem}

This theorem can be applied locally when optimizing a general MLP (\eqref{e:f}) with SAL to prove local plane reproduction. See supplementary for more details. 
%
% \begin{theorem}\label{thm:planar_reproduction_mlp}
% Consider an MLP as defined in \eqref{e:f}. Assume that locally in some domain $\Omega\subset\Real^d$ the data $\gX\cap \Omega$ lies on a plane $\gP=\set{\vx\vert \vn^T\vx+c=0}$, \ie, $\gX\cap\Omega\subset \gP$. Then, there exists a critical point $\vtheta^*$ of the loss in  \eqref{e:loss} that reconstructs $\gP$ \emph{locally} in $\Omega$.
% \end{theorem}
% By "reconstructs $\gP$ locally" we mean that $\vtheta^*$ is critical for the loss if $D_\gX$ is sufficiently concentrated around any point in $\gP\cap\Omega$. 

\begin{figure}[t]
     \includegraphics[width=\columnwidth]{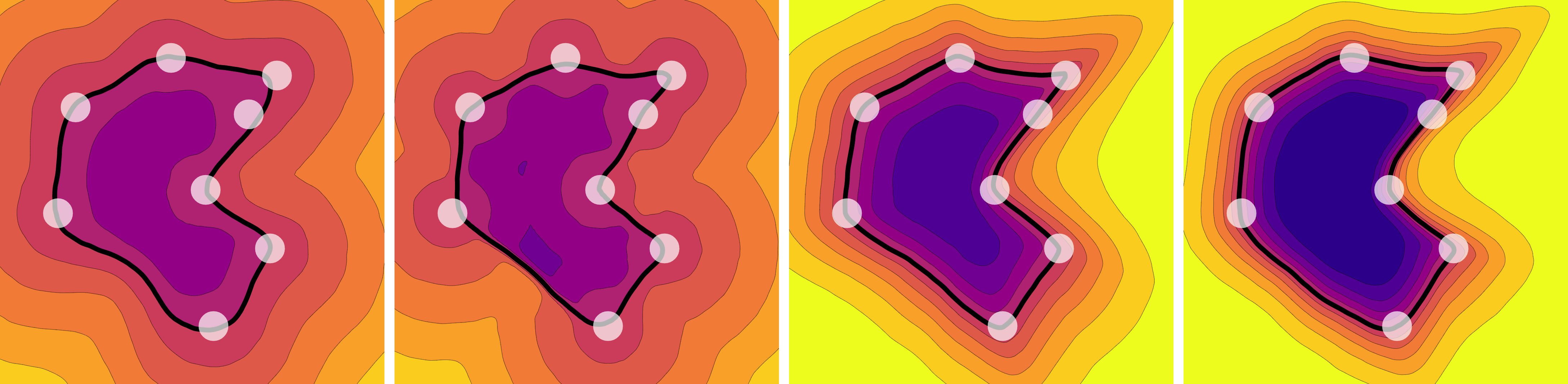}
    \caption{Advanced epochs of the neural level-sets from Figure \ref{fig:2d}. The limit in the $L^0$ case (two right images) is an inside/outside indicator function, while for the $L^2$ case (two left images) it is a signed version of the unsigned $L^2$ distance. }%\vspace{-10pt}
    \label{fig:limit}
\end{figure}

\subsection{Convergence to the limit signed function}
%The plane reproduction in Theorem \ref{thm:planar_reproduction} can only by applied locally with a full MLP model. For MLP, 
%
The SAL loss pushes the neural implicit function $f$ towards a signed version of the unsigned distance function $h_\gX$. In the $L^0$ case it is the inside/outside indicator function of the surface, while for $L^2$ it is a signed version of the Euclidean distance to the data $\gX$. Figure \ref{fig:limit} shows advanced epochs of the 2D experiment in Figure \ref{fig:2d}; note that the $f$ in these advanced epochs is indeed closer to the signed version of the respective $h_\gX$. 
Since the indicator function and the \emph{signed} Euclidean distance are discontinuous across the surface, they potentially impose quantization errors when using standard contouring algorithms, such as Marching Cubes \cite{lorensen1987marching}, to extract their zero level-set. In practice, this phenomenon is avoided with a standard choice of stopping criteria (learning rate and number of iterations). Another potential solution is to add a regularization term to the SAL loss; we mark this as future work.

\section{Experiments}

\begin{figure}[t]
     \includegraphics[width=\columnwidth]{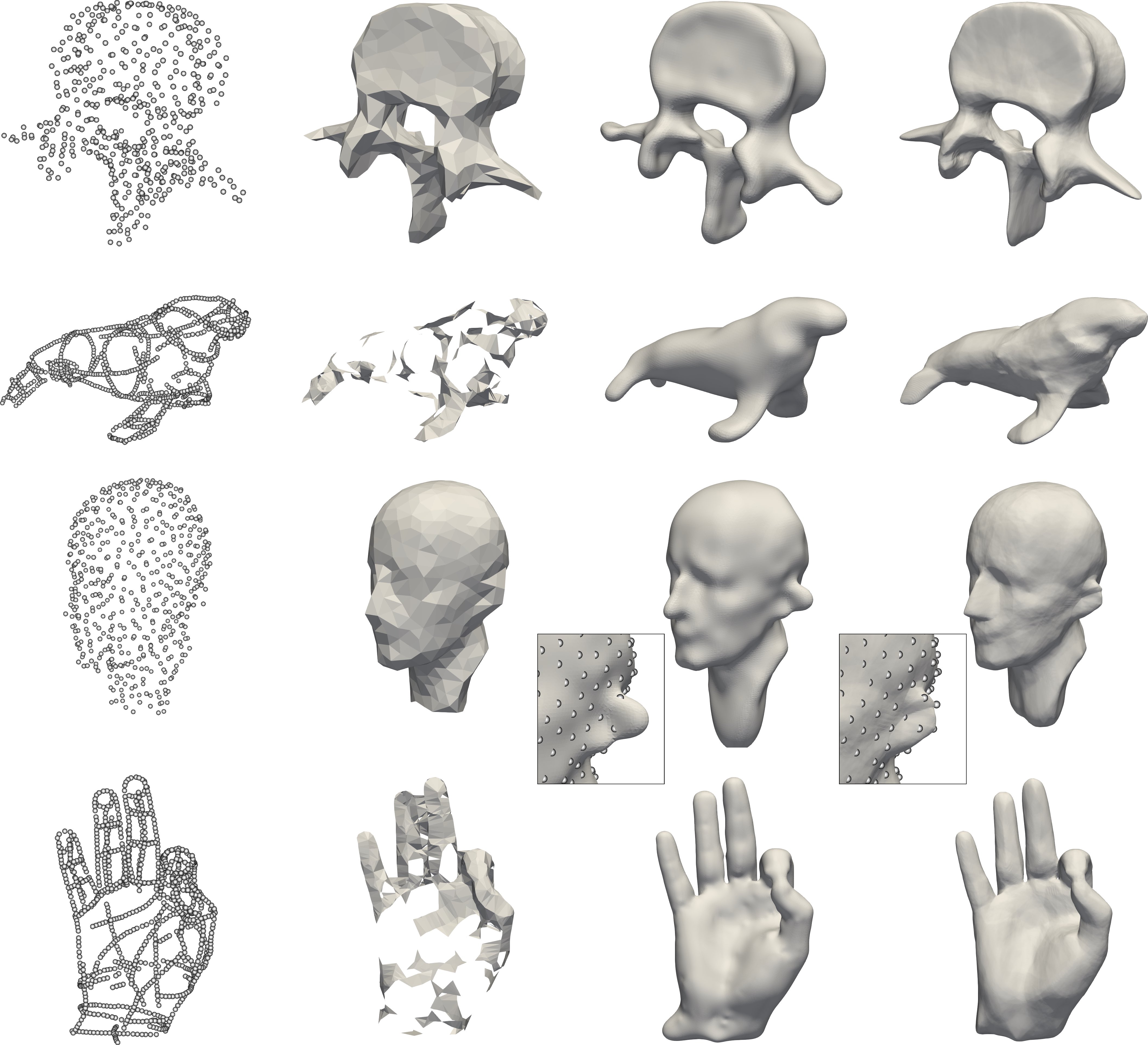}
    \caption{Surface reconstruction from (un-oriented) point cloud. From left to right: input point cloud; ball-pivoting reconstruction \cite{bernardini1999ball}; variational-implicit reconstruction \cite{Zhiyang2019}; SAL reconstruction (ours). }
    \label{fig:recon}
\end{figure}

\subsection{Surface reconstruction}\label{ss:recon}
The most basic experiment for SAL is reconstructing a surface from a single input raw point cloud (without using any normal information). Figure \ref{fig:recon} shows surface reconstructions based on four raw point clouds provided in \cite{Zhiyang2019} with three methods: ball-pivoting \cite{bernardini1999ball}, variation-implicit reconstruction \cite{Zhiyang2019}, and SAL based on the $L^0$ distance, \ie, optimizing the loss described in \eqref{e:loss_L0} with $\ell=1$. The only parameter in this loss is $\sigma$ which we set for every point in $\vx\in\gX$ to be the distance to the $50$-th nearest point in the point cloud $\gX$. We used an 8-layer MLP, $f:\Real^3\times \Real^m\too\Real$, with 512 wide hidden layers and a single skip connection to the middle layer (see supplementary material for more implementation details). As can be visually inspected from the figure, SAL provides high fidelity surfaces, approximating the input point cloud even for challenging cases of sparse and irregular input point clouds.

\subsection{Learning shape space from raw scans} 
In the main experiment of this paper we trained on the D-Faust scan dataset \cite{dfaust:CVPR:2017}, consisting of approximately 41k  raw scans of 10 humans in multiple poses\footnote{Due to the dense temporal sampling in this dataset we experimented with a 1:5 sample.}. Each scan is a triangle soup, $\gX_i$, where common defects include holes, ghost geometry, and noise, see Figure \ref{fig:teaser} for examples. 

\paragraph{Architecture.}
To learn the shape representations we used a modified variational encoder-decoder \cite{kingma2013auto}, where the encoder $(\vmu,\veta)=g(\mX;\vtheta_1)$ is taken to be PointNet \cite{qi2017pointnet} (specific architecture detailed in supplementary material), $\mX\in\Real^{n\times 3}$ is an input point cloud (we used $n=128^2$), $\vmu\in\Real^{256}$ is the latent vector, and $\veta\in\Real^{256}$ represents a diagonal covariance matrix by $\vSigma=\diag \exp{\veta}$. That is, the encoder takes in a point cloud $\mX$ and outputs a probability measure $\gN(\vmu,\vSigma)$. The point cloud is drawn uniformly at random from the scans, \ie, $\mX \sim \gX_i$. The decoder is the implicit representation $f(\vx ; \vw , \vtheta_2)$ with the addition of a latent vector $\vw\in\Real^{256}$. The architecture of $f$ is taken to be the 8-layer MLP, as in Subsection \ref{ss:recon}. 

\paragraph{Loss.}
We use SAL loss with $L^2$ distance, \ie, $h_2(\vz)=\min_{\vx\in \gX_i} \norm{\vz-\vx}_2$ the unsigned distance to the triangle soup $\gX_i$, and combine it with a variational auto-encoder type loss \cite{kingma2013auto}:
% \begin{equation}
%     \E_{\vw \sim \gN(\vmu,\vSigma)}f(\vx;\vw,\vtheta_2)
% \end{equation}
\begin{align*}
  \mathrm{Loss}(\vtheta) &=  \sum_{i} \E_{\mX \sim \gX_i} \Big [ \recloss(\vtheta) + \lambda \norm{\vmu}_1 + \norm{\veta + \one}_1  \Big ] \\ 
  \recloss(\vtheta) & =\E_{\vz\sim \gN_\sigma(\gX_i),\vw \sim \gN(\vmu,\vSigma)} \big| |f(\vz;\vw,\vtheta_2)| - h_2(\vz) \big|,
\end{align*}
where $\vtheta=(\vtheta_1,\vtheta_2)$, $\norm{\cdot}_1$ is the 1-norm, $\norm{\vmu}_1$ encourages the latent prediction $\vmu$ to be close to the origin, while $\norm{\veta+\one}_1$ encourages the variances $\vSigma$ to be constant $\exp{(-1)}$; together, these enforce a regularization on the latent space. $\lambda$ is a balancing weight chosen to be $10^{-3}$.   

%% how we prepare data:
% preprocess: 250k uniformly from the mesh
% 500k points based on 50th and 0.2 gaussian from initial sample. compute distance to $\gX$ from all these points. 

% batch: random 128^2 from 250k -> pointnet. 128^2 from 500k with unsigned distances to mesh. 

\begin{figure*}[t]
     \includegraphics[width=\textwidth]{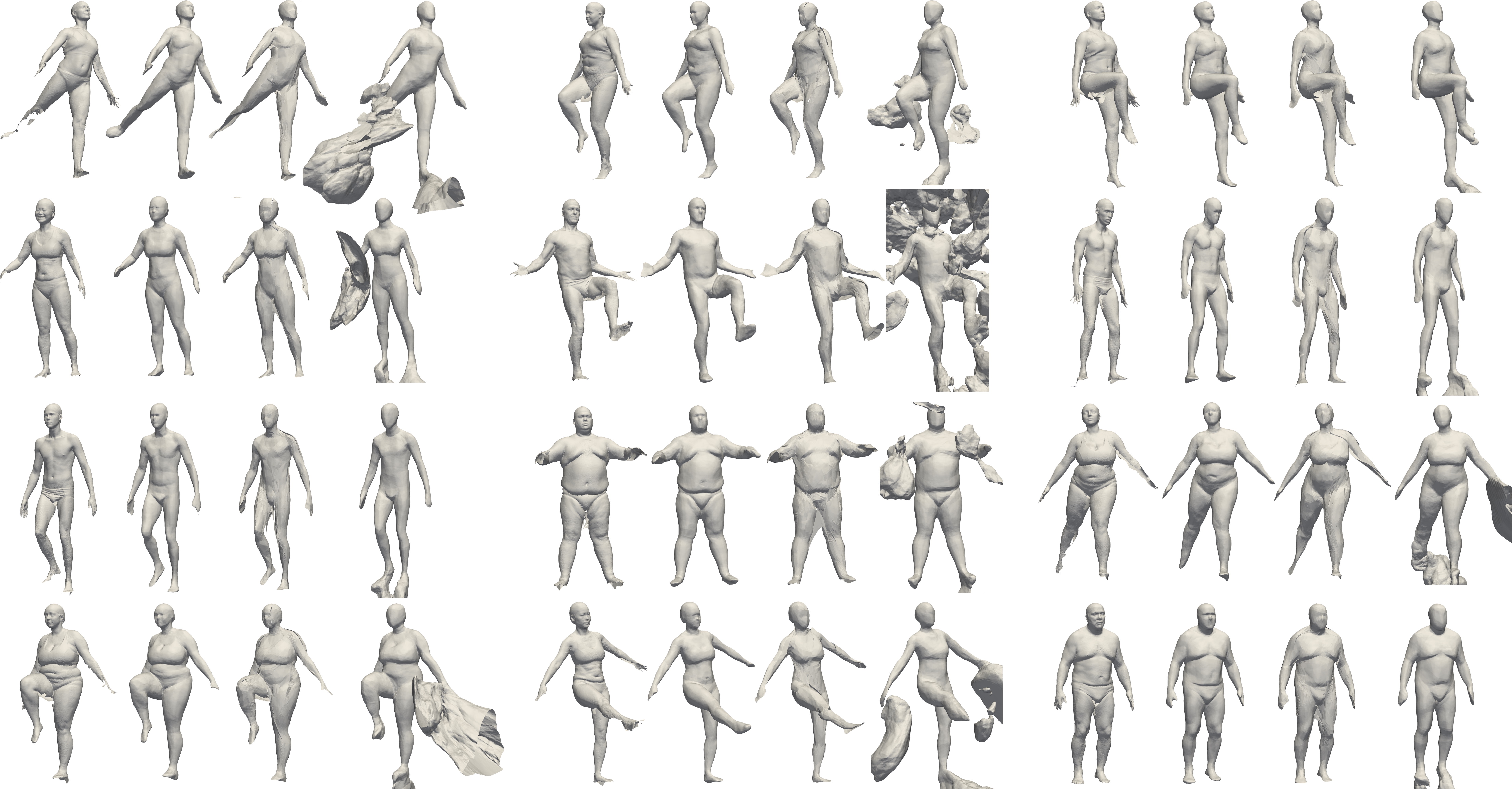}
    \caption{Reconstruction of the test set from D-Faust scans. Left to right in each column: input test scan, SAL (our) reconstruction, AtlasNet \cite{groueix2018papier} reconstruction, and SignReg - signed regression with approximate Jet normals. }
    \label{fig:easy_test}
\end{figure*}

\paragraph{Baseline.}
We compared versus three baseline methods. First, AtlasNet \cite{groueix2018papier}, one of the only existing algorithms for learning a shape collection from raw point clouds. AtlasNet uses a parametric representation of surfaces, which is straight-forward to sample. On the down side, it uses a collection of patches that tend to not overlap perfectly, and their loss requires computation of closest points between the generated and input point clouds which poses a challenge for learning large point clouds. Second, we approximate a signed distance function, $\bar{h}_2$, to the data $\gX_i$ in two different ways, and regress them using an MLP as in DeepSDF \cite{Park_2019_CVPR}; we call these methods SignReg. Note that Occupancy Networks \cite{mescheder2019occupancy} and \cite{chen2019learning} regress a different signed distance function and perform similarly. 

To approximate the signed distance function, $\bar{h}_2$, we first tried using a state of the art surface reconstruction algorithm \cite{kazhdan2013screened} to produce watertight manifold surfaces. However, only 28684 shapes were successfully reconstructed ($69\%$ of the dataset), making this option infeasible to compute $\bar{h}_2$. We have opted to approximate the signed distance function similar to \cite{hoppe1992surface} with $\bar{h}_2(\vz) = \vn_*^T(\vz- \vx_*)$, where $\vx_*=\argmin_{\vx\in \gX_i}\norm{\vz-\vx}_2$ is the closest point to $\vz$ in $\gX_i$ and $\vn_*$ is the normal at $\vx_*\in\gX_i$. To approximate the normal $\vn_*$ we tested two options: (i) taking $\vn^*$ directly from the original scan $\gX_i$ with its original orientation; and (ii) using local normal estimation using Jets \cite{cazals2005estimating} followed by consistent orientation procedure based on minimal spanning tree using the CGAL library \cite{cgal:ass-psp-19b}. 

Table \ref{tab:easy_test} and Figure \ref{fig:easy_test} show the result on a random 75\%-25\% train-test split on the D-Faust raw scans. We report the 5\%, 50\% (median), and 95\% percentiles of the Chamfer distances between the surface reconstructions and the raw scans (one-sided Chamfer from reconstruction to scan), and ground truth registrations. The SAL and SignReg reconstructions were generated by a forward pass $(\vmu,\veta)=g(\mX;\vtheta_1)$ of a point cloud $\mX \subset \gX_i$ sampled from the raw unseen scans, yielding an implicit function $f(\vx;\vmu,\vtheta_2)$. We used the Marching Cubes algorithm \cite{lorensen1987marching} to mesh the zero level-set of this implicit function. Then, we sampled uniformly 30K points from it and compute the Chamfer Distance.
%what split? 

\begin{table}[t]
    \centering
    \scriptsize
    \setlength\tabcolsep{2pt} % default value: 6pt
    \begin{tabular}{c}
        \begin{adjustbox}{max width=\textwidth}
            \aboverulesep=0ex
            \belowrulesep=0ex
            \renewcommand{\arraystretch}{1.1}
            \begin{tabular}[t]{c|c|c|c|c|c | c | c|}
            \multicolumn{2}{c}{} & 
            \multicolumn{3}{|c|}{Registrations} & 
            \multicolumn{3}{|c}{Scans} \\
                \cmidrule{2-8}
                & Method                                  &
                5\% & Median & 95\%  & 5\% & Median & 95\% \\
                \midrule
                \multirow{ 4}{*}{Train} &
                AtlasNet\cite{groueix2018papier} & 0.09 & 0.15 & 0.27 & \textbf{0.05} & 0.09 & 0.18 \\
                & Scan normals & 2.53 & 43.99 & 292.59 & 2.63 & 44.86 & 257.37 \\
                & Jet normals & 1.72 & 30.46 & 513.34 & 1.65 & 31.11 & 453.43 \\
                & SAL (ours) & \textbf{0.05} & \textbf{0.09} & \textbf{0.2} & \textbf{0.05} & \textbf{0.06} & \textbf{0.09} \\
                 \midrule
                 \multirow{ 4}{*}{Test} &
                 AtlasNet\cite{groueix2018papier} & 0.1 & 0.17 & 0.37 & \textbf{0.05} & 0.1 & 0.22 \\
                & Scan normals & 3.45 & 45.03 & 294.15 & 3.21 & 277.36 & 45.03 \\
                & Jet normals & 1.88 & 31.05 & 489.35 & 1.76 & 30.89  & 462.85 \\
                & SAL (ours) & \textbf{0.07} & \textbf{0.12} & \textbf{0.35} & \textbf{0.05} & \textbf{0.08} & \textbf{0.16} \\
                \cmidrule{2-8}
                %\bottomrule
        \end{tabular} 
        \end{adjustbox}
      
    \end{tabular}
    %\vspace{3pt}
    \caption{Reconstruction of the test set from D-Faust scans.  We log the Chamfer distances of the reconstructed surfaces to the raw scans (one-sided), and ground-truth registrations; we report the 5-th, 50-th, and 95-th percentile. Numbers are reported $* 10^3$.} %\vspace{-15pt}
    \label{tab:easy_test}
\end{table}

\paragraph{Generalization to unseen data.}
In this experiment we test our method on two different scenarios: (i) generating shapes of unseen humans; and (ii) generating shapes of unseen poses. For the unseen humans experiment we trained on $8$ humans ($4$ females and $4$ males), leaving out $2$ humans for test (one female and one male). For the unseen poses experiment, we randomly chose two poses of each human as a test set. To further improve test-time shape representations, we also further optimized the latent $\vmu$ to better approximate the input test scan $\gX_i$.  That is, for each test scan $\gX_i$, after the forward pass $(\vmu,\veta)=g(\mX;\vtheta_2)$ with $\mX\subset \gX_i$, we further optimized $\recloss$ as a function of $\vmu$ for 800 further iterations. We refer to this method as latent optimization. 

\begin{figure}[t]
     \includegraphics[width=\columnwidth]{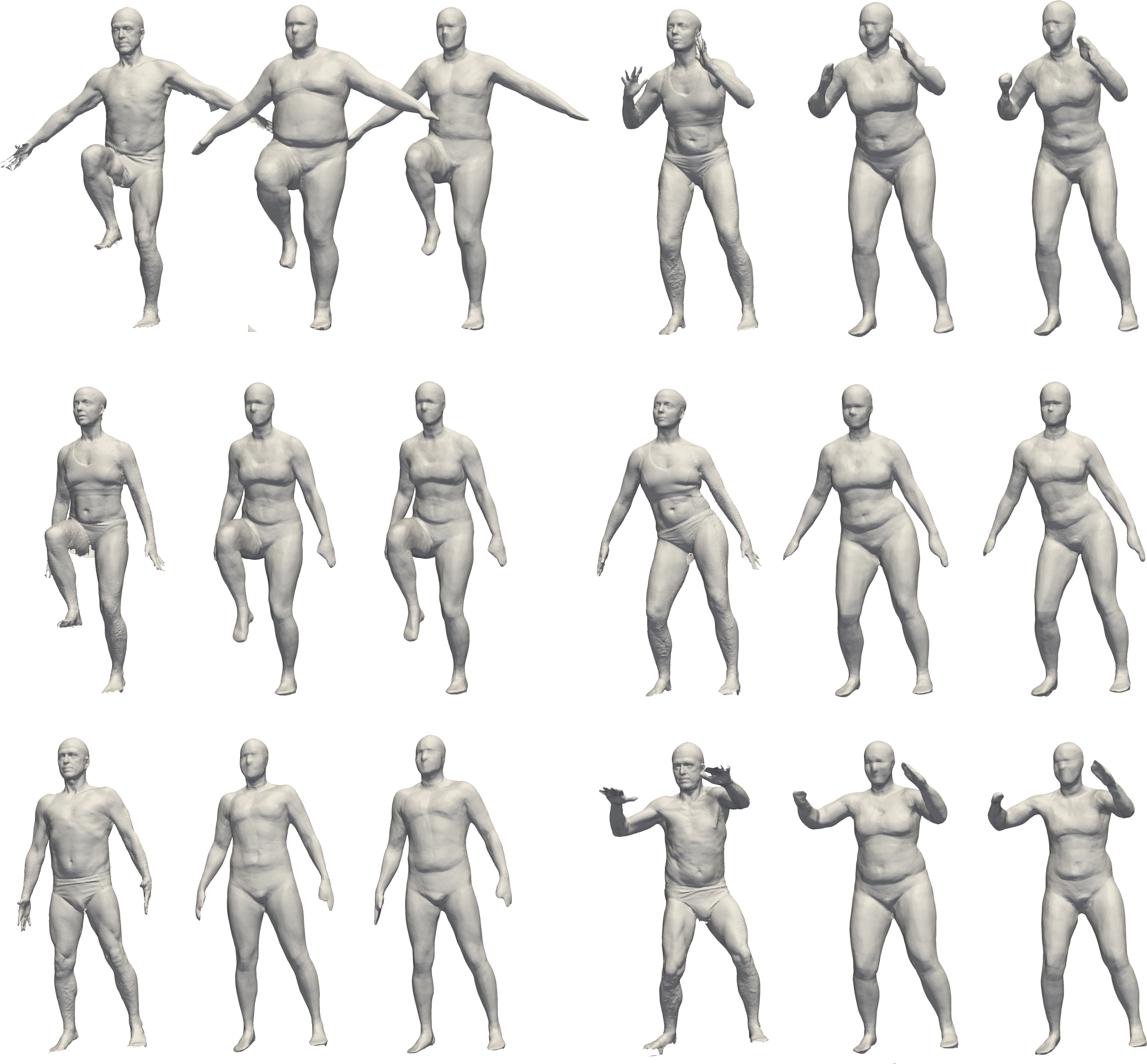}
    \caption{Reconstruction of unseen humans scans. Each column from left to right: unseen human scan, SAL reconstruction with a single forward pass, SAL reconstruction with latent optimization. Bottom-right shows failure.}
    \label{fig:human_test}
\end{figure}

\begin{figure}[t]
     \includegraphics[width=\columnwidth]{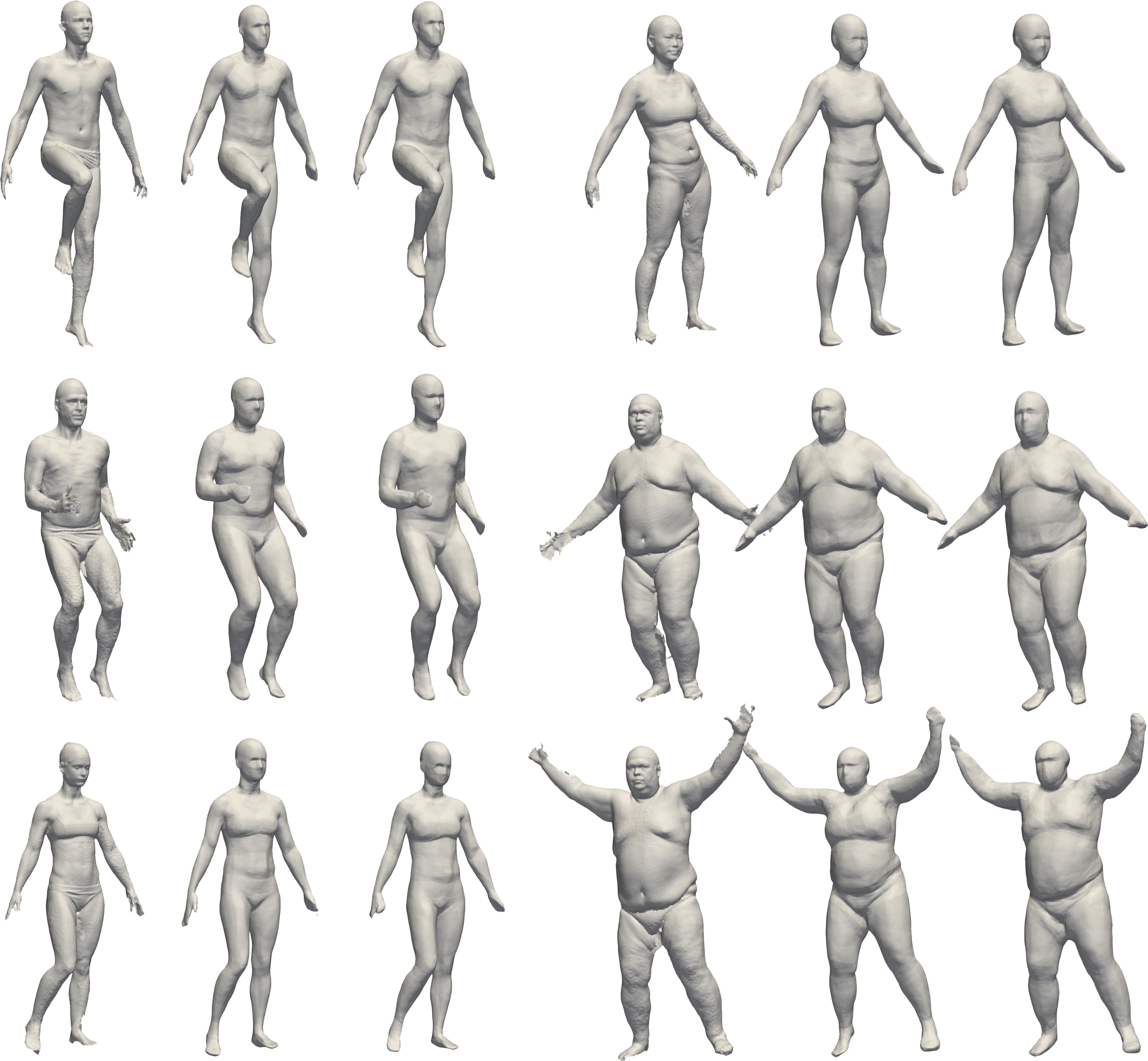}
    \caption{Reconstruction of unseen pose scans. Each column from left to right: unseen pose scan, SAL reconstruction with a single forward pass, SAL reconstruction with latent optimization. Bottom-right shows failure. \vspace{-10pt} }\label{fig:pose_test}
\end{figure}

Table \ref{tab:dfaust_results} demonstrates that the latent optimization method further improves predictions quality, compared to a single forward pass. In \ref{fig:human_test} and \ref{fig:pose_test}, we demonstrate few representatives examples, where we plot left to right in each column: input test scan, SAL reconstruction with forward pass alone, and SAL reconstruction with latent optimization. Failure cases are shown in the bottom-right. Despite the little variability of humans in the training dataset (only 8 humans), \ref{fig:human_test} shows that SAL can usually fit a pretty good human shape to the unseen human scan using a single forward pass reconstruction; using latent optimization further improves the approximation as can be inspected in the different examples in this figure. 

Figure \ref{fig:pose_test} shows how a single forward reconstruction is able to predict the pose correctly, where latent optimization improves the prediction in terms of shape and pose. 

\begin{table}[t]
    \centering
    \scriptsize
    \setlength\tabcolsep{2pt} % default value: 6pt
    \begin{tabular}{c}
        \begin{adjustbox}{max width=\textwidth}
            \aboverulesep=0ex
            \belowrulesep=0ex
            \renewcommand{\arraystretch}{1.1}
            \begin{tabular}[t]{c|c|c|c|c|c|c|c|}
                \multicolumn{2}{c}{} & 
            \multicolumn{3}{|c|}{Registrations} & 
            \multicolumn{3}{|c}{Scans} \\
                \cmidrule{2-8}
                & Method &
                5\% & Median & 95\%  & 5\% & Median & 95\% \\
                \midrule
                \multirow{ 2}{*}{Train} & SAL (Pose) & 0.08 & 0.12 & 0.25 & 0.05 & 0.07 & 0.1 \\
                & SAL (Human) & 0.06 & 0.09 & 0.18 & 0.04 & 0.06 & 0.09 \\
                 \midrule
                \multirow{ 4}{*}{Test} & 
                 SAL (Pose) & 0.11 & 0.37 & 2.26 & 0.07 & 0.18 & 0.93 \\
                & SAL + latent opt. (Pose) & 0.08 & 0.16 & 1.12 & 0.05 & 0.09 & 0.19 \\
                & SAL (Human) & 0.26 & 0.75 & 4.99 & 0.14 & 0.34 & 1.53 \\
                & SAL + latent opt. (Human) & 0.12 & 0.3 & 3.05 & 0.07 & 0.14 & 0.49 \\
                \cmidrule{2-8}
        \end{tabular} 
        \end{adjustbox}
      
    \end{tabular}
    %\vspace{3pt}
    \caption{Reconstruction of the unseen human and pose from D-Faust scans. We log the Chamfer distances of the reconstructed surfaces to the raw scans (one-sided), and ground-truth registrations; we report the 5-th, 50-th, and 95-th percentile. Numbers are reported $* 10^3$.} \vspace{5pt}
    \label{tab:dfaust_results}
\end{table} 

\begin{figure}[t]
     \includegraphics[width=\columnwidth]{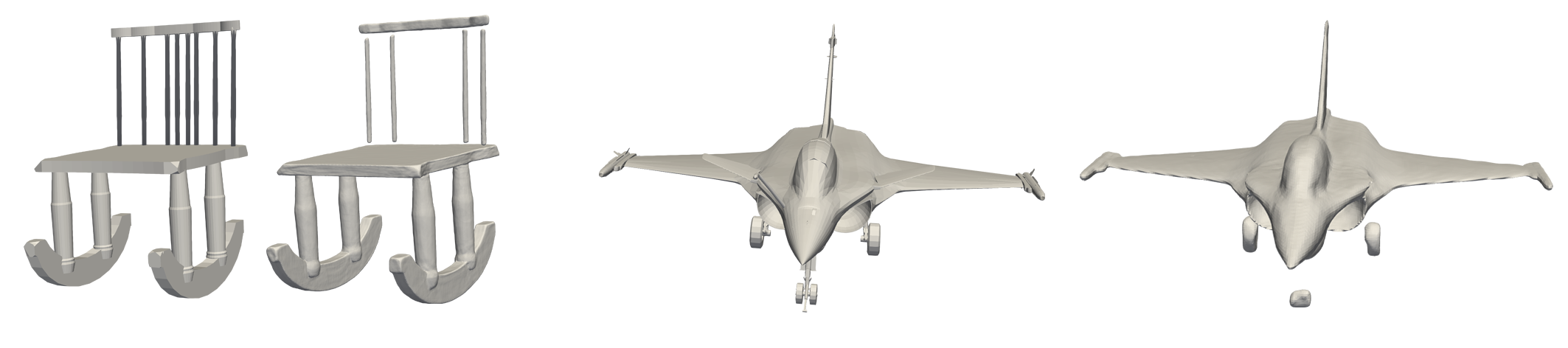}
    \caption{Failure in capturing thin structures. In each pair: ground truth model (left), and SAL reconstruction (right).}\label{fig:limitation}
\end{figure}

\paragraph{Limitations.}
SAL's limitation is mainly in capturing thin structures. Figure \ref{fig:limitation} shows reconstructions (obtained similarly to \ref{ss:recon}) of a chair and a plane from the ShapeNet \cite{chang2015shapenet} dataset; note that some parts in the chair back and the plane wheel structure are missing.

\section {Conclusions}
We introduced SAL: Sign Agnostic Learning, a deep learning approach for processing raw data without any preprocess or need for ground truth normal data or inside/outside labeling. We have developed a  geometric initialization formula for MLPs to approximate the signed distance function to a sphere, and a theoretical justification proving planar reproduction for SAL. Lastly, we demonstrated the ability of SAL to reconstruct high fidelity surfaces from raw point clouds, and that SAL easily integrates into standard generative models to learn shape spaces from raw geometric data. One limitation of SAL was mentioned in Section \ref{s:properties}, namely the stopping criteria for the optimization. 

Using SAL in other generative models such as generative adversarial networks could be an interesting follow-up. Another future direction is global reconstruction from partial data. Combining SAL with image data also has potentially interesting applications. We think SAL has many exciting future work directions, progressing geometric deep learning to work with unorganized, raw data. 

\subsection*{Acknowledgments}
The research was supported by the European Research Council (ERC Consolidator Grant, "LiftMatch" 771136), the Israel Science Foundation (Grant No. 1830/17) and by a research grant from the Carolito Stiftung (WAIC).

{\small
\bibliographystyle{ieee_fullname}
\bibliography{arxiv}

\begin{thebibliography}{10}\itemsep=-1pt

\bibitem{cgal:ass-psp-19b}
Pierre Alliez, Simon Giraudot, Cl{\'e}ment Jamin, Florent Lafarge, Quentin
  M{\'e}rigot, Jocelyn Meyron, Laurent Saboret, Nader Salman, and Shihao Wu.
\newblock Point set processing.
\newblock In {\em {CGAL} User and Reference Manual}. {CGAL Editorial Board},
  {5.0} edition, 2019.

\bibitem{arora2016understanding}
Raman Arora, Amitabh Basu, Poorya Mianjy, and Anirbit Mukherjee.
\newblock Understanding deep neural networks with rectified linear units.
\newblock {\em arXiv preprint arXiv:1611.01491}, 2016.

\bibitem{atzmon2019controlling}
Matan Atzmon, Niv Haim, Lior Yariv, Ofer Israelov, Haggai Maron, and Yaron
  Lipman.
\newblock Controlling neural level sets.
\newblock {\em arXiv preprint arXiv:1905.11911}, 2019.

\bibitem{bagautdinov2018modeling}
Timur Bagautdinov, Chenglei Wu, Jason Saragih, Pascal Fua, and Yaser Sheikh.
\newblock Modeling facial geometry using compositional vaes.
\newblock In {\em Proceedings of the IEEE Conference on Computer Vision and
  Pattern Recognition}, pages 3877--3886, 2018.

\bibitem{ben2018multi}
Heli Ben-Hamu, Haggai Maron, Itay Kezurer, Gal Avineri, and Yaron Lipman.
\newblock Multi-chart generative surface modeling.
\newblock In {\em SIGGRAPH Asia 2018 Technical Papers}, page 215. ACM, 2018.

\bibitem{berger2017survey}
Matthew Berger, Andrea Tagliasacchi, Lee~M Seversky, Pierre Alliez, Gael
  Guennebaud, Joshua~A Levine, Andrei Sharf, and Claudio~T Silva.
\newblock A survey of surface reconstruction from point clouds.
\newblock In {\em Computer Graphics Forum}, volume~36, pages 301--329. Wiley
  Online Library, 2017.

\bibitem{bernardini1999ball}
Fausto Bernardini, Joshua Mittleman, Holly Rushmeier, Cl{\'a}udio Silva, and
  Gabriel Taubin.
\newblock The ball-pivoting algorithm for surface reconstruction.
\newblock {\em IEEE transactions on visualization and computer graphics},
  5(4):349--359, 1999.

\bibitem{dfaust:CVPR:2017}
Federica Bogo, Javier Romero, Gerard Pons-Moll, and Michael~J. Black.
\newblock Dynamic {FAUST}: {R}egistering human bodies in motion.
\newblock In {\em IEEE Conf. on Computer Vision and Pattern Recognition
  (CVPR)}, July 2017.

\bibitem{bojanowski2017optimizing}
Piotr Bojanowski, Armand Joulin, David Lopez-Paz, and Arthur Szlam.
\newblock Optimizing the latent space of generative networks.
\newblock {\em arXiv preprint arXiv:1707.05776}, 2017.

\bibitem{carr2001reconstruction}
Jonathan~C Carr, Richard~K Beatson, Jon~B Cherrie, Tim~J Mitchell, W~Richard
  Fright, Bruce~C McCallum, and Tim~R Evans.
\newblock Reconstruction and representation of 3d objects with radial basis
  functions.
\newblock In {\em Proceedings of the 28th annual conference on Computer
  graphics and interactive techniques}, pages 67--76. ACM, 2001.

\bibitem{cazals2005estimating}
Fr{\'e}d{\'e}ric Cazals and Marc Pouget.
\newblock Estimating differential quantities using polynomial fitting of
  osculating jets.
\newblock {\em Computer Aided Geometric Design}, 22(2):121--146, 2005.

\bibitem{chang2015shapenet}
Angel~X Chang, Thomas Funkhouser, Leonidas Guibas, Pat Hanrahan, Qixing Huang,
  Zimo Li, Silvio Savarese, Manolis Savva, Shuran Song, Hao Su, et~al.
\newblock Shapenet: An information-rich 3d model repository.
\newblock {\em arXiv preprint arXiv:1512.03012}, 2015.

\bibitem{chen2019learning}
Zhiqin Chen and Hao Zhang.
\newblock Learning implicit fields for generative shape modeling.
\newblock In {\em Proceedings of the IEEE Conference on Computer Vision and
  Pattern Recognition}, pages 5939--5948, 2019.

\bibitem{dai2017shape}
Angela Dai, Charles Ruizhongtai~Qi, and Matthias Nie{\ss}ner.
\newblock Shape completion using 3d-encoder-predictor cnns and shape synthesis.
\newblock In {\em Proceedings of the IEEE Conference on Computer Vision and
  Pattern Recognition}, pages 5868--5877, 2017.

\bibitem{deng2019cvxnets}
Boyang Deng, Kyle Genova, Soroosh Yazdani, Sofien Bouaziz, Geoffrey Hinton, and
  Andrea Tagliasacchi.
\newblock Cvxnets: Learnable convex decomposition.
\newblock {\em arXiv preprint arXiv:1909.05736}, 2019.

\bibitem{deprelle2019learning}
Theo Deprelle, Thibault Groueix, Matthew Fisher, Vladimir~G Kim, Bryan~C
  Russell, and Mathieu Aubry.
\newblock Learning elementary structures for 3d shape generation and matching.
\newblock {\em arXiv preprint arXiv:1908.04725}, 2019.

\bibitem{do2016differential}
Manfredo~P Do~Carmo.
\newblock {\em Differential Geometry of Curves and Surfaces: Revised and
  Updated Second Edition}.
\newblock Courier Dover Publications, 2016.

\bibitem{genova2019learning}
Kyle Genova, Forrester Cole, Daniel Vlasic, Aaron Sarna, William~T Freeman, and
  Thomas Funkhouser.
\newblock Learning shape templates with structured implicit functions.
\newblock {\em arXiv preprint arXiv:1904.06447}, 2019.

\bibitem{goodfellow2014generative}
Ian Goodfellow, Jean Pouget-Abadie, Mehdi Mirza, Bing Xu, David Warde-Farley,
  Sherjil Ozair, Aaron Courville, and Yoshua Bengio.
\newblock Generative adversarial nets.
\newblock In {\em Advances in neural information processing systems}, pages
  2672--2680, 2014.

\bibitem{groueix2018papier}
Thibault Groueix, Matthew Fisher, Vladimir~G Kim, Bryan~C Russell, and Mathieu
  Aubry.
\newblock A papier-m{\^a}ch{\'e} approach to learning 3d surface generation.
\newblock In {\em Proceedings of the IEEE conference on computer vision and
  pattern recognition}, pages 216--224, 2018.

\bibitem{hoppe1992surface}
Hugues Hoppe, Tony DeRose, Tom Duchamp, John McDonald, and Werner Stuetzle.
\newblock {\em Surface reconstruction from unorganized points}, volume~26.
\newblock ACM, 1992.

\bibitem{Zhiyang2019}
Zhiyang Huang, Nathan Carr, and Tao Ju.
\newblock Variational implicit point set surfaces.
\newblock {\em ACM Trans. Graph.}, 38(4), July 2019.

\bibitem{kazhdan2006poisson}
Michael Kazhdan, Matthew Bolitho, and Hugues Hoppe.
\newblock Poisson surface reconstruction.
\newblock In {\em Proceedings of the fourth Eurographics symposium on Geometry
  processing}, volume~7, 2006.

\bibitem{kazhdan2013screened}
Michael Kazhdan and Hugues Hoppe.
\newblock Screened poisson surface reconstruction.
\newblock {\em ACM Transactions on Graphics (ToG)}, 32(3):29, 2013.

\bibitem{kingma2014adam}
Diederik~P Kingma and Jimmy Ba.
\newblock Adam: A method for stochastic optimization.
\newblock {\em arXiv preprint arXiv:1412.6980}, 2014.

\bibitem{kingma2013auto}
Diederik~P Kingma and Max Welling.
\newblock Auto-encoding variational bayes.
\newblock {\em arXiv preprint arXiv:1312.6114}, 2013.

\bibitem{litany2018deformable}
Or Litany, Alex Bronstein, Michael Bronstein, and Ameesh Makadia.
\newblock Deformable shape completion with graph convolutional autoencoders.
\newblock In {\em Proceedings of the IEEE Conference on Computer Vision and
  Pattern Recognition}, pages 1886--1895, 2018.

\bibitem{lorensen1987marching}
William~E Lorensen and Harvey~E Cline.
\newblock Marching cubes: A high resolution 3d surface construction algorithm.
\newblock In {\em ACM siggraph computer graphics}, volume~21, pages 163--169.
  ACM, 1987.

\bibitem{maron2017convolutional}
Haggai Maron, Meirav Galun, Noam Aigerman, Miri Trope, Nadav Dym, Ersin Yumer,
  Vladimir~G Kim, and Yaron Lipman.
\newblock Convolutional neural networks on surfaces via seamless toric covers.
\newblock {\em ACM Trans. Graph.}, 36(4):71--1, 2017.

\bibitem{mescheder2019occupancy}
Lars Mescheder, Michael Oechsle, Michael Niemeyer, Sebastian Nowozin, and
  Andreas Geiger.
\newblock Occupancy networks: Learning 3d reconstruction in function space.
\newblock In {\em Proceedings of the IEEE Conference on Computer Vision and
  Pattern Recognition}, pages 4460--4470, 2019.

\bibitem{mullen2010signing}
Patrick Mullen, Fernando De~Goes, Mathieu Desbrun, David Cohen-Steiner, and
  Pierre Alliez.
\newblock Signing the unsigned: Robust surface reconstruction from raw
  pointsets.
\newblock In {\em Computer Graphics Forum}, volume~29, pages 1733--1741. Wiley
  Online Library, 2010.

\bibitem{Park_2019_CVPR}
Jeong~Joon Park, Peter Florence, Julian Straub, Richard Newcombe, and Steven
  Lovegrove.
\newblock Deepsdf: Learning continuous signed distance functions for shape
  representation.
\newblock In {\em The IEEE Conference on Computer Vision and Pattern
  Recognition (CVPR)}, June 2019.

\bibitem{paszke2017automatic}
Adam Paszke, Sam Gross, Soumith Chintala, Gregory Chanan, Edward Yang, Zachary
  DeVito, Zeming Lin, Alban Desmaison, Luca Antiga, and Adam Lerer.
\newblock Automatic differentiation in pytorch.
\newblock 2017.

\bibitem{qi2017pointnet}
Charles~R Qi, Hao Su, Kaichun Mo, and Leonidas~J Guibas.
\newblock Pointnet: Deep learning on point sets for 3d classification and
  segmentation.
\newblock In {\em Proceedings of the IEEE Conference on Computer Vision and
  Pattern Recognition}, pages 652--660, 2017.

\bibitem{sinha2016deep}
Ayan Sinha, Jing Bai, and Karthik Ramani.
\newblock Deep learning 3d shape surfaces using geometry images.
\newblock In {\em European Conference on Computer Vision}, pages 223--240.
  Springer, 2016.

\bibitem{sinha2017surfnet}
Ayan Sinha, Asim Unmesh, Qixing Huang, and Karthik Ramani.
\newblock Surfnet: Generating 3d shape surfaces using deep residual networks.
\newblock In {\em Proceedings of the IEEE conference on computer vision and
  pattern recognition}, pages 6040--6049, 2017.

\bibitem{stutz2018learning}
David Stutz and Andreas Geiger.
\newblock Learning 3d shape completion from laser scan data with weak
  supervision.
\newblock In {\em Proceedings of the IEEE Conference on Computer Vision and
  Pattern Recognition}, pages 1955--1964, 2018.

\bibitem{takayama2014consistently}
Kenshi Takayama, Alec Jacobson, Ladislav Kavan, and Olga Sorkine-Hornung.
\newblock Consistently orienting facets in polygon meshes by minimizing the
  dirichlet energy of generalized winding numbers.
\newblock {\em arXiv preprint arXiv:1406.5431}, 2014.

\bibitem{tatarchenko2017octree}
Maxim Tatarchenko, Alexey Dosovitskiy, and Thomas Brox.
\newblock Octree generating networks: Efficient convolutional architectures for
  high-resolution 3d outputs.
\newblock In {\em Proceedings of the IEEE International Conference on Computer
  Vision}, pages 2088--2096, 2017.

\bibitem{walder2005implicit}
Christian Walder, Olivier Chapelle, and Bernhard Sch{\"o}lkopf.
\newblock Implicit surface modelling as an eigenvalue problem.
\newblock In {\em Proceedings of the 22nd international conference on Machine
  learning}, pages 936--939. ACM, 2005.

\bibitem{walder2007implicit}
Christian Walder, Olivier Chapelle, and Bernhard Sch{\"o}lkopf.
\newblock Implicit surfaces with globally regularised and compactly supported
  basis functions.
\newblock In {\em Advances in Neural Information Processing Systems}, pages
  273--280, 2007.

\bibitem{wendland2004scattered}
Holger Wendland.
\newblock {\em Scattered data approximation}, volume~17.
\newblock Cambridge university press, 2004.

\bibitem{williams2019deep}
Francis Williams, Teseo Schneider, Claudio Silva, Denis Zorin, Joan Bruna, and
  Daniele Panozzo.
\newblock Deep geometric prior for surface reconstruction.
\newblock In {\em Proceedings of the IEEE Conference on Computer Vision and
  Pattern Recognition}, pages 10130--10139, 2019.

\bibitem{wu2016learning}
Jiajun Wu, Chengkai Zhang, Tianfan Xue, Bill Freeman, and Josh Tenenbaum.
\newblock Learning a probabilistic latent space of object shapes via 3d
  generative-adversarial modeling.
\newblock In {\em Advances in neural information processing systems}, pages
  82--90, 2016.

\bibitem{xu2014signed}
Hongyi Xu and Jernej Barbi{\v{c}}.
\newblock Signed distance fields for polygon soup meshes.
\newblock In {\em Proceedings of Graphics Interface 2014}, pages 35--41.
  Canadian Information Processing Society, 2014.

\bibitem{zaheer2017deep}
Manzil Zaheer, Satwik Kottur, Siamak Ravanbakhsh, Barnabas Poczos, Ruslan~R
  Salakhutdinov, and Alexander~J Smola.
\newblock Deep sets.
\newblock In {\em Advances in neural information processing systems}, pages
  3391--3401, 2017.

\bibitem{zhao2001fast}
Hong-Kai Zhao, Stanley Osher, and Ronald Fedkiw.
\newblock Fast surface reconstruction using the level set method.
\newblock In {\em Proceedings IEEE Workshop on Variational and Level Set
  Methods in Computer Vision}, pages 194--201. IEEE, 2001.

\end{thebibliography}
}

\section{Appendix}
\subsection{Implementation details}
\subsubsection{Surface reconstruction}

\paragraph{Data Preparation.}
For each point cloud $\gX\subset \Real^3$ we created training data by sampling two Gaussian variables centered at each $\vx\in\gX$. 
We set the standard deviation of the first Gaussian to be the distance to the 50-th closest point in $\gX$, whereas the second Gaussian standard deviation is set to be the distance to the furthest point in $\gX$.

%   X = ((torch.randn(N, S.shape[0], d_in)*(sigmas.unsqueeze(0).unsqueeze(2).repeat(1,1,3)) +
%               S.unsqueeze(0).repeat(N, 1, 1)).reshape(N*S.shape[0], d_in))
%         X_general = ( torch.randn(N, S.shape[0], d_in)*sigma_max + S.mean(dim=0).repeat(N,S.shape[0],1)).squeeze()
%         X = torch.cat([X,X_general],0)

\paragraph{Network Architecture.}
We use an MLP as detailed in \eqref{e:f} and \eqref{e:f_params} with $\ell=7$, \ie, 8 layers, where $d_i^{out},d_i^{in}=512$ for interior layers and $d_1^{in}=3$. We also add a skip connection at the 4-th layer concatenating the input point $\vx\in\Real^3$ with the hidden variable $\vy\in\Real^{509}$, \ie, $[\vx,\vy]\in\Real^{512}$, where accordingly $d_3^{out}=509$.  

\paragraph{Training Details.}
We train the network in the surface reconstruction experiment with \textsc{Adam} optimizer \cite{kingma2014adam}, learning rate $0.0001$ for 5000 epochs. Training was done on an Nvidia V-100 GPU, using \textsc{pytorch} deep learning framework \cite{paszke2017automatic}.

\subsubsection{Learning shape space}

\paragraph{Data Preparation.}
To speed up training on the D-Faust dataset, we preprocessed the original scans $\gX\subset\Real^3$ and sampled 500K points from each scan. The size of the sample is similar to \cite{Park_2019_CVPR}. First, we sampled 250K points uniformly (w.r.t.~area) from the triangle soup $\gX$ and then placed two Gaussian random variables centered at each sample point $\vx\in\gX$. We set the standard deviation of the first Gaussian to be the distance to the 50-th closest point in $\gX$, whereas the second Gaussian standard deviation is set to be $0.2$. For each sample point we calculated its unsigned distance to the closest triangle in the triangle soup. The distance computation was done using \cite{cgal:ass-psp-19b}.

\paragraph{Network Architecture.}
Our network architecture is Encoder-Decoder based, where for the encoder we have used PointNet \cite{qi2017pointnet} and DeepSets \cite{zaheer2017deep} layers. Each layer is composed of 
\begin{align*}
\mathrm{PFC}(d_{\text{in}},d_{\text{out}}):\mX &\mapsto \nu\parr{\mX W + \one b^T }  \\
\mathrm{PL}(d_{\text{in}},2d_{\text{in}}):\mY & \mapsto \brac{\mY, \max{(\mY)} \one}\\
\end{align*}
where $\brac{\cdot,\cdot}$ is the concat operation. Our Architecture is 
\begin{align*}
&\mathrm{PFC}(3,128) \rightarrow \mathrm{PFC}(128,128) \rightarrow \mathrm{PL}(128,256) \rightarrow\\ 
& \mathrm{PFC}(256,128)  \rightarrow \mathrm{PL}(128,256) \rightarrow \mathrm{PFC}(256,128)  \rightarrow \\
&  \mathrm{PL}(128,256) \rightarrow \mathrm{PFC}(256,128)  \rightarrow \mathrm{PL}(128,256) \rightarrow \\
& \mathrm{PFC}(256,256)  \rightarrow \mathrm{MaxPool} \rightarrow  \mathrm{FC}(256,256), 
\end{align*}
similarly to \cite{mescheder2019occupancy} SimplePointNet architecture. For the decoder, our network architecture is the same as in the Surface reconstruction experiment except for $d_1^{in}=256 + 3$, as the decoder receives as input $ \brac{z,x} \in \Real^{259} $. Where $\brac{z,x}$ is a concatenation of the latent encoding $\vz\in\Real^{256}$ and a point $\vx\in\Real^{3}$ in space.
%composed of $8$ MLP layers. See equation \ref{e:f_params} in the paper for an exact definition of a MLP layer.
% \begin{align*}
% & \brac{3,256} \rightarrow \mathrm{FC}(259,512) \rightarrow \mathrm{FC}(512,512) \rightarrow\\ 
% & \mathrm{FC}(512,512)  \rightarrow \mathrm{FC}(512,253) \rightarrow \brac{259,253}  \rightarrow \\
% &  \mathrm{FC}(512,512) \rightarrow \mathrm{FC}(512,512)  \rightarrow \mathrm{FC}(512,512) \rightarrow \\
% & \mathrm{FC}(512,512)  \rightarrow \mathrm{Linear}(512,1), 
% \end{align*}
% where the first $\brac{\cdot,\cdot}$ is a concatenation between the latent encoding and the input point in space. The second $\brac{\cdot,\cdot}$ is a skip connection, concatenating the output of the previous layer and the output of the first concatenation. Our decoder architecture is similar to \cite{Park_2019_CVPR}.

\paragraph{Training Details.}
Training our networks for learning the shape space of the D-Faust dataset was done with the following choices. We have used the \textsc{Adam} optimizer \cite{kingma2014adam}, initialized with learning rate $0.0005$ and batch size of $64$. We scheduled the learning rate to decrease every 500 epochs by a factor of 0.5. We stopped the training process after 2000 epochs. Training was done on 4 Nvidia V-100 GPUs, using \textsc{pytorch} deep learning framework \cite{paszke2017automatic}.

\begin{figure*}[t!]
     \includegraphics[width=0.9\textwidth]{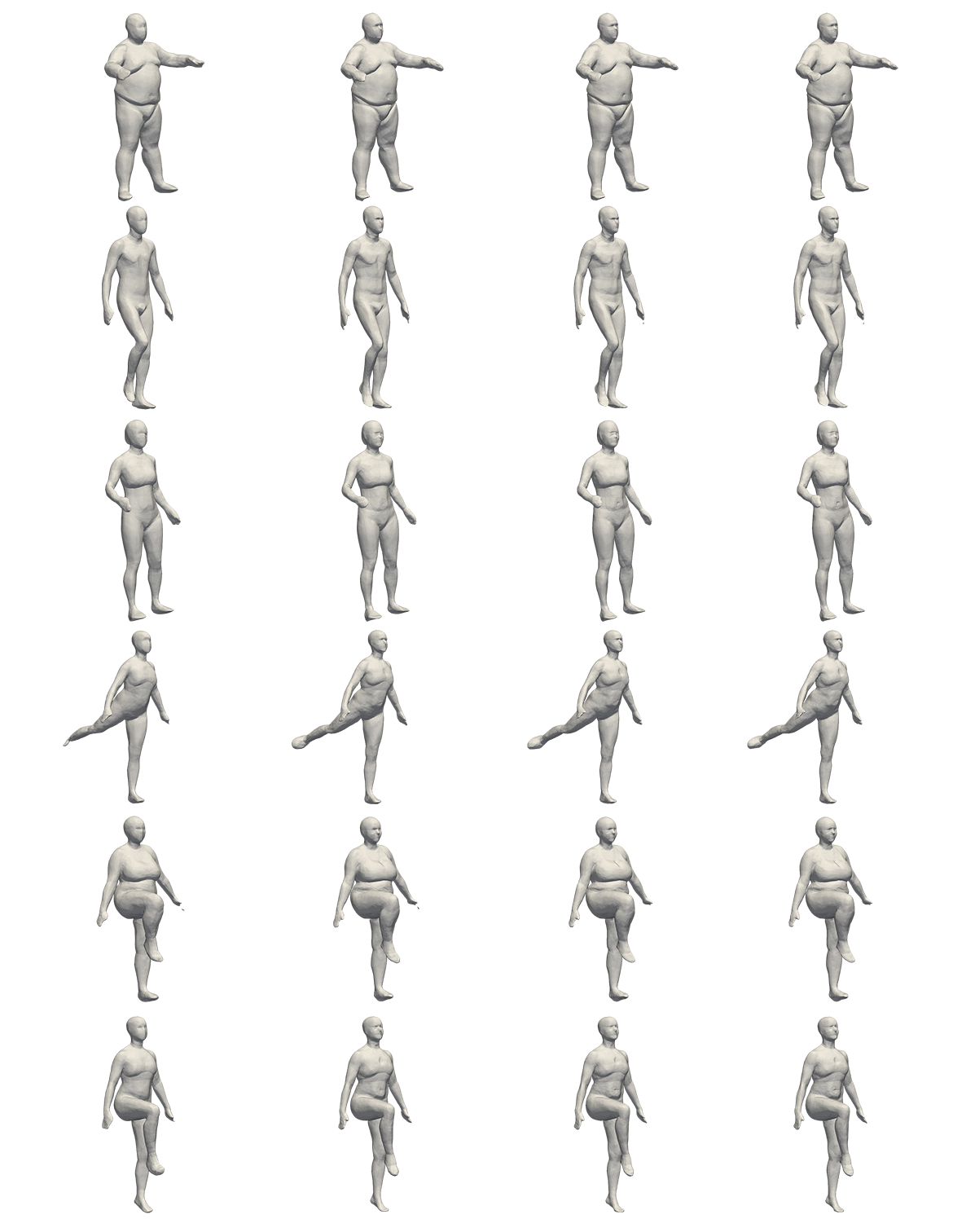}
    \caption{Effect of the number of training epochs on reconstruction quality of test scans. Left to right in each row: epoch $500$, $1500$, $2500$ and $3500$.}
    \label{fig:easy_by_epochs}
    
\end{figure*}

\subsection{Additional Experiments}
\paragraph{Single reconstruction versus VAE reconstruction.}
One of the key advantages of SAL is that it can be used for reconstructing a surface from a single input scan or incorporated into a VAE architecture for learning a shape space from an entire scans dataset. This raises an interesting question, whether learning a shape space has also an impact on the quality of the reconstructions.  To answer this question, we ran SAL surface reconstruction on each of the scans used for training the main experiment of the paper (See table \ref{tab:dfaust_results} for more details). When comparing our SAL VAE training results on the registrations (ground truth) versus SAL single reconstruction we see differences in favor of our VAE learner, whereas the results on the original scans are comparable. That is, SAL single reconstruction results are $0.10, 0.17, 0.22;0.07,0.08,0.10$ on the registrations and scans for the $5\%,50\%,95\%$ percentiles respectively.
\paragraph{Number of epochs used for training SAL VAE.}
Figure \ref{fig:easy_by_epochs} shows reconstructions of test scans for different stages of training on the D-Faust dataset. Given the main paper discussion on SAL limit signed function, we additionally add reconstructions from relatively advanced epoch as $3500$, showing that no error in contouring occur. 

\subsection{Proofs}

\subsubsection{Proof of Theorem \ref{thm:planar_reproduction}}

\begin{reptheorem}{thm:planar_reproduction}
Consider a linear model  $f(\vx;\vtheta)=\varphi(\vw^T\vx+b)$, $\vtheta=(\vw,b)$, with a strong non-linearity $\varphi:\Real\too\Real$. Assume the data $\gX$ lies on a plane $\gP=\set{\vx\vert \vn^T\vx+c=0}$, \ie, $\gX\subset \gP$. Then, there exists $\alpha\in \Real_+$ so that $(\vw^*,b^*)=(\alpha\vn, \alpha c)$ is a critical point of the loss in \eqref{e:loss}. 
\end{reptheorem}

\begin{proof}
For simplicity, we restrict our attention to absolutely continuous measures $D_\gX$, that is defined by a continuous density function $\mu(\vx)$. Generalizing to measures with a discrete part (such as the one we use for the $L^0$ distance, for example) can be proven similarly.

Denoting $\vtheta=(\vw,b)$, the loss in \eqref{e:loss} can be written as $$\loss(\vw,b) = \int_{\Real^d} \tau(f(\vx;\vtheta),h_\gX(\vx)) \mu(\vx)d\vx. $$

Denote by $\vr:\Real^d\too\Real^d$ the linear reflection w.r.t.~$\gP$, \ie, $\vr(\vx)=\vx - 2 \vn\vn^T(\vx-\vx_0)=(I-2\vn\vn^T)\vx-2c\vn $, where $\vx_0\in\gP$ is an arbitrary point. $h_\gX$ and $\mu$ are invariant to $\vr$, that is $h_\gX(\vr(\vx))=h_\gX(\vx)$, $\mu(\vr(\vx))=\mu(\vx)$.

The gradient of the loss is $\nabla_{\vw,b} \loss(\vw,b) = $
\begin{equation}\label{e:grad}
\int_{\Real^d}\frac{\partial \tau}{\partial a}(f(\vx;\vtheta),h_\gX(\vx)) \nabla_{\vw,b} f(\vx;\vtheta) \mu(\vx)d\vx,
\end{equation}
where $\nabla_{\vw,b}f(\vx;\vtheta)=\varphi'(\vw^T\vx+b) [\vx,1]$.
%\begin{equation}\label{e:grad}
%\int_{\Real^d}\frac{\partial \tau}{\partial a}(f(\vx;\vtheta),h_\gX(\vx)) \varphi'(\vw^T\vx+b) [\vx,1] \mu(\vx)d\vx.
%\end{equation}

Let $\vw=\alpha\vn$ and $b=\alpha c$, where $\alpha\in \Real_+$ is currently arbitrary. We decompose $\Real^d$ to the two sides of $\gP$, $\Omega_+ = \set{\vx\vert \vn^T \vx + c \geq 0}$, and $\Omega_- = \Real^d\setminus \Omega_+$. Now consider the integral in \eqref{e:grad} restricted to $\Omega_-$ and perform change of variables $\vx = \vr(\vy)$; note that $\vr$ consists of an orthogonal linear part and therefore $d\vy = \abs{\det \frac{\partial\vy}{\partial \vx}}d\vx = d\vx$. Furthermore, property (\ref{item:sym}) implies that $\frac{\partial \tau}{\partial a}(a,b) = -\frac{\partial \tau}{\partial a}(-a,b)$, and since $\varphi$ is anti-symmetric and $\vw^T \vr(\vx)+b=-(\vw^T\vy +b)$ we get $$\frac{\partial \tau}{\partial a}(f(\vr(\vy);\vtheta),h_\gX(\vr(\vy)))=-\frac{\partial \tau}{\partial a}(f(\vy;\vtheta),h_\gX(\vy)).$$ As $\varphi$ is anti-symmetric, $\varphi'$ is symmetric, \ie, $\varphi'(a)=\varphi'(-a)$ and therefore 
$$\varphi'(\vw^T \vr(\vy) + b) = \varphi'(\vw^T\vy + b). $$ Plugging these in the integral after the change of variables we reach
$$-\int_{\Omega_+}\frac{\partial \tau}{\partial a}(f(\vy;\vtheta),h_\gX(\vy)) \varphi'(\vw^T\vy+b) [\vr(\vy),1] \mu(\vy)d\vy. $$
An immediate consequence is that $\nabla_b \loss (\vw,b)=0$. As for $\nabla_\vw \loss(\vw,b)$ we have:
$$\int_{\Omega_+}\frac{\partial \tau}{\partial a}(f(\vx;\vtheta),h_\gX(\vx)) \varphi'(\vw^T\vx+b)(\vx-\vr(\vx))\mu(\vx)d\vx.$$
Since $\vx-\vr(\vx) = 2\vn\vn^T(\vx-\vx_0)$ we get $\nabla_\vw \loss(\vw,b)=$
$$2\vn \int_{\Omega_+}\frac{\partial \tau}{\partial a}(f(\vx;\vtheta),h_\gX(\vx)) \varphi'(\vw^T\vx+b)\vn^T(\vx-\vx_0)\mu(\vx)d\vx.$$
The last integral is scalar and we denote its integrand by $g_\alpha(\vx)$ (remember that $(\vw,b)=\alpha(\vn,c)$), \ie,
$$\nabla_\vw \loss(\vw,b) = 2\vn\int_{\Omega_+}g_\alpha(\vx)d\vx=2\vn G(\alpha),$$ where $G(\alpha)=\int_{\Omega_+} g_\alpha(\vx)d\vx$. The proof will be done if we show there exists $\alpha\in\Real_+$ so that $G(\alpha)=0$. We will use the intermediate value theorem for continuous functions to prove this, that is, we will show existence of $\alpha_-,\alpha_+\in\Real_+$ so that $G(\alpha_-)<0<G(\alpha_+)$. 

Let us show the existence of $\alpha_+$.  Let $\vx\in \Omega_+$ such that $\gamma=\vn^T\vx+c>0$ and $\mu(\vx)>0$. Then,
$$g_\alpha(\vx) =\frac{\partial \tau}{\partial a}(\varphi(\alpha \gamma),h_\gX(\vx)) \varphi'(\alpha \gamma)\gamma\mu(\vx).$$
Furthermore, since $\beta^{-1}\geq \varphi'(a)\geq \beta >0$ we have that
\begin{align*}
g_\alpha(\vx) &\geq \frac{\partial \tau}{\partial a}(\varphi(\alpha \gamma),h_\gX(\vx))\gamma\mu(\vx)\begin{cases}\beta & \frac{\partial\tau}{\partial a}\geq 0 \\ \beta^{-1} & \frac{\partial \tau}{\partial a} < 0 \end{cases}\\ &=\tilde{g}_\alpha(\vx).
\end{align*}
Note that $\tilde{g}_\alpha(\vx)$ is monotonically increasing and for sufficiently large $\alpha\in\Real_+$ we have that $\tilde{g}_\alpha(\vx)>0$. Since $\int_{\Omega_+}\tilde{g}_\alpha(\vx)d\vx > -\infty$ for all $\alpha\in\Real_+$ the integral monotone convergence theorem implies that $\lim_{\alpha\too \infty} \int_{\Omega_+}\tilde{g}_\alpha(\vx)d\vx > 0$. Lastly, since $$\int_{\Omega_+}g_\alpha(\vx)d\vx \geq \int_{\Omega_+}\tilde{g}_\alpha(\vx)d\vx,$$ the existence of $\alpha_+$ is established. The case of $\alpha_-$ is proven similarly. 
\end{proof}

\subsubsection{Local plane reproduction with MLP}

\begin{theorem}\label{thm:planar_reproduction_mlp}
Consider an MLP as defined in \eqref{e:f}. Assume that locally in some domain $\Omega\subset\Real^d$ the data $\gX\cap \Omega$ lies on a plane $\gP=\set{\vx\vert \vn^T\vx+c=0}$, \ie, $\gX\cap\Omega\subset \gP$. Then, there exists a critical point $\vtheta^*$ of the loss in  \eqref{e:loss} that reconstructs $\gP$ \emph{locally} in $\Omega$.
\end{theorem}
By "reconstructs $\gP$ locally" we mean that $\vtheta^*$ is critical for the loss if $D_\gX$ is sufficiently concentrated around any point in $\gP\cap\Omega$.

\begin{proof}
We next consider a general MLP model $f(\vx;\vtheta)$ as in \eqref{e:f}. We denote by $\supp(\mu)$ the support set of $\mu$, \ie, $\set{\vx \vert \mu(\vx)>0}$.  
Let us write the layers of the network $f(\vx;\vtheta)$ using only matrix multiplication, that is
\begin{equation}\label{e:mlp_mat_mul}
f_i(\vy) = \diag\big(H(\mW_i\vy+\vb_i)\big) \big(\mW_i\vy+\vb_i \big),
\end{equation}
where $H(a)={\tiny \begin{cases} 1 & a\geq 0 \\ 0 & a<0 \end{cases}}$ is the Heaviside function. 

Let $\mu$ be so that $\supp(\mu)\subset \Omega$, and fix an arbitrary $\vx_0\in \supp(\mu)\cap \gP$. Next, let $\vtheta\in\Real^m$ be such that: (a) there exists a domain $\Upsilon$, so that $\supp(\mu)\subset \Upsilon$ and for which all the diagonal matrices in \eqref{e:mlp_mat_mul} are constants, \ie, $\Upsilon$ is a domain over which $f(\vx;\vtheta)$ (excluding the final non-linearity $\varphi$) is linear as a function of $\vx$. Therefore, for $\vx\in\Upsilon$ we have the layers of $f$ satisfy
$$f_i(\vy) = D_i \big(\mW_i\vy+\vb_i \big),$$ and $D_i$ are constant diagonal matrices. Over this domain we have 
$$f(\vx;\vtheta) = \varphi\big( \vw(\vtheta)^T \vx + b(\vtheta)  \big),$$
and therefore
\begin{equation}\label{e_proof:nabla_theta}
\nabla_\vtheta f(\vx;\vtheta) =  \mA \nabla_{\vw,b} f(\vx;\vtheta) ,
\end{equation}
where $\nabla_{\vw,b}f(\vx;\vtheta)=\varphi'(\vw(\vtheta)^T\vx+b(\vtheta))[\vx,1]$, and $\mA\in\Real^{m \times (d+1)}$ is a matrix holding the partial derivatives of $\vw(\vtheta)$ and $b(\vtheta)$.  (b) over $\Upsilon$ there exists $\vw(\vtheta)=\vn$ and $b(\vtheta)=c$. 
The existence of such $\vtheta$ is guaranteed for example from Theorem 2.1 in \cite{arora2016understanding}.

Similarly to the proof of Theorem \ref{thm:planar_reproduction} there exists $\alpha\in\Real_+$ so that $\nabla_{\vw,b} \loss(\alpha \vw(\vtheta),\alpha b(\vtheta))=0$. Scaling the $\vw,b$ components of $\vtheta$ by $\alpha$ and denoting the new parameter vector by $\vtheta^*$, we get that $\nabla_{\vw,b} \loss( \vw(\vtheta^*),b(\vtheta^*))=0$ (see \eqref{e:grad}) and therefore 
\eqref{e_proof:nabla_theta} implies that $\nabla_\vtheta \loss(\vtheta^*)=0$, as required. 
\end{proof}

\subsubsection{Proof of Theorem \ref{cor:geometric_init}}

\begin{reptheorem}{cor:geometric_init}
Let $f$ be an MLP (see equations \ref{e:f}-\ref{e:f_params}). Set, for $1\leq i \leq \ell$, $\vb_i=0$ and $\mW_i$ i.i.d.~from a normal distribution $\gN(0,\frac{\sqrt{2}}{\sqrt{d_i^{out}}})$; further set $\vw={\frac{\sqrt{\pi}}{\sqrt{d_\ell^{out}}}}\one$, $c=-r$. Then, $f(\vx)\approx \varphi(\norm{x}-r)$.
\end{reptheorem}
\begin{proof}
To prove this theorem reduce the problem to a single hidden layer network, see Theorem \ref{thm:3_layers_sphere}.
Denote $g(\vx) = f_{\ell-1}\circ \cdots \circ f_1$. If we prove that $\norm{g(\vx)}\approx\norm{\vx}$ then Theorem \ref{thm:3_layers_sphere} implies the current theorem. 

%\yl{check if this analysis is similar to he or other NN initialization techniques...} 
It is enough to consider a single layer: $h(\vx) = \nu(\mW\vx+ \vb)$. For brevity let $k=d^{out}$. Now, 
$$\norm{h(\vx)}^2 = \sum_{i=1}^{k} \nu(\mW_{i,:}\cdot \vx)^2= \frac{1}{k}\sum_{i=1}^{k}\nu(\sqrt{k}\mW_{i,:}\cdot \vx)^2.$$
Note that the entries of $\sqrt{k\mW_{i,:}}$ are distributed i.i.d.~$\gN(0,\sqrt{2})$. Hence by the law of large numbers the last term converge to 
\begin{align*}
    &\norm{\vx}^2\int_{\Real^k} \nu \parr{\vy \cdot \frac{\vx}{\norm{\vx}} }^2\mu(\vy)dy \\ &=  \norm{\vx}^2 \int_{\Real^k} \nu(y_1)^2 \frac{1}{(2\pi \sigma^2)^{k/2}}e^{-\frac{\norm{\vy}}{2\sigma^2}} dy \\&
    = \norm{\vx}^2 \int_{\Real} \nu(y_1)^2 \frac{1}{\sqrt{2\pi \sigma^2}}e^{-\frac{y_1^2}{2\sigma^2}} dy \\&
    = \frac{\norm{\vx}^2}{2} \int_{\Real} y_1^2 \frac{1}{\sqrt{2\pi \sigma^2}}e^{-\frac{y_1^2}{2\sigma^2}} dy \\ &=\norm{\vx}^2,
\end{align*}
where in the second equality, similarly to the proof of Theorem \ref{thm:3_layers_sphere}, we changed variables, $\vy=\mR \vy'$, where we chose $\mR\in\Real^{k\times k}$ orthogonal so that $\mR^T\frac{\vx}{\norm{\vx}}=(1,0,\ldots,0)^T$. 
\end{proof}

\paragraph{Skip connections.} Adapting the geometric initialization to skip connections is easy: consider skip connection layers of the form $s(\vy)=\frac{1}{\sqrt{2}}(\vy,\vx)$, where $\vx\in\Real^{d}$ is the input to the network and $\vy\in\Real^{d^{out}_i}$ is some interior hidden variable. Then $\norm{s(\vy)}^2= \frac{1}{2}\norm{\vx}^2 + \frac{1}{2}\norm{\vy}^2$. According to Theorem \ref{cor:geometric_init} we have $\norm{\vy}\approx \norm{\vx}$ and hence $\norm{s(\vy)}^2\approx \norm{\vx}^2$.  

\subsubsection{Proof of Theorem \ref{thm:3_layers_sphere}}
\begin{reptheorem}{thm:3_layers_sphere}
Let $f:\Real^{d}\too \Real$ be an MLP with ReLU activation, $\nu$, and a single hidden layer. That is, $f(\vx) = \vw^T\nu(\mW \vx+ \vb)+c$, where $\mW\in\Real^{d^{out}\times d}$, $\vb\in\Real^{d^{out}}$, $\vw\in\Real^{d^{out}}$, $c\in\Real$ are the learnable parameters. If $\vb=0$, $\vw=\frac{\sqrt{2\pi}}{\sigma d^{out}}\one$, $c=-r$, $r>0$, and all entries of $\mW$ are i.i.d.~normal $\gN(0,\sigma^2)$ then $f(\vx)\approx \norm{\vx}-r$. That is, $f$ is approximately the signed distance function to a $d-1$ sphere of radius $r$ in $\Real^{d}$, centered at the origin.  
\end{reptheorem}
\begin{proof} For brevity we denote $k=d^{out}$. Note that plugging $\vw, \vb, c$ in $f$ we get $f(\vx)=\frac{\sqrt{2\pi}}{\sigma k}\sum_{i=1}^k \nu(\vw_{i}\cdot\vx) -r $, where $\vw_{i}$ is the $i^{th}$ row of $\mW$. Let $\mu$ denote the density of multivariate normal distribution $\gN=(0,\sigma^2I_k)$. By the law of large numbers, the first term converges to 
% \begin{align*}
%     &\frac{\sqrt{2\pi}}{\sigma }\int_{\Real^k} \nu \parr{\vu\cdot \frac{\vx}{\norm{\vx}} \norm{\vx}} \mu(\vu) d\vu \\ &= \frac{\sqrt{2\pi}\norm{\vx}}{\sigma } \int_{\Real^k} \nu \parr{\vu\cdot \frac{\vx}{\norm{\vx}} }\mu(\vu)d\vu \\ & = \frac{\sqrt{2\pi}\norm{\vx}}{\sigma }\int_{\Real^k} \nu(v_1) \mu(\vv) d\vv \\ &
%      = \frac{\sqrt{2\pi}\norm{\vx}}{\sigma }\int_{\Real^k} \nu(v_1)\frac{1}{(2\pi\sigma^2)^{k/2}} e^{-\frac{\norm{\vv}^2}{2\sigma^2}}d\vv \\ & = \frac{\sqrt{2\pi}\norm{\vx}}{\sigma }\int_\Real \nu(v_1) \frac{1}{\sqrt{2\pi \sigma^2}}e^{-\frac{v_1^2}{2\sigma^2}}dv_1 \\ &= \frac{\sqrt{2\pi}\norm{\vx}}{2\sigma } \int_\Real \abs{v_1}\frac{1}{\sqrt{2\pi \sigma^2}}e^{-\frac{v_1^2}{2\sigma^2}}dv_1 \\& = \norm{\vx},
% \end{align*}
\begin{align*}
    &\frac{\sqrt{2\pi}}{\sigma }\int_{\Real^k} \nu \parr{\vu\cdot \frac{\vx}{\norm{\vx}} \norm{\vx}} \mu(\vu) d\vu \\ &= \frac{\sqrt{2\pi}\norm{\vx}}{\sigma } \int_{\Real^k} \nu \parr{\vu\cdot \frac{\vx}{\norm{\vx}} }\mu(\vu)d\vu \\ & = \frac{\sqrt{2\pi}\norm{\vx}}{\sigma }\int_{\Real^k} \nu(v_1) \mu(\vv) d\vv \\ &
     = \frac{\sqrt{2\pi}\norm{\vx}}{\sigma }\int_{\Real^k} \nu(v_1)\frac{1}{(2\pi\sigma^2)^{k/2}} e^{-\frac{\norm{\vv}^2}{2\sigma^2}}d\vv 
\end{align*}
\begin{align*}
     & = \frac{\sqrt{2\pi}\norm{\vx}}{\sigma }\int_\Real \nu(v_1) \frac{1}{\sqrt{2\pi \sigma^2}}e^{-\frac{v_1^2}{2\sigma^2}}dv_1 \\ &= \frac{\sqrt{2\pi}\norm{\vx}}{2\sigma } \int_\Real \abs{v_1}\frac{1}{\sqrt{2\pi \sigma^2}}e^{-\frac{v_1^2}{2\sigma^2}}dv_1 \\& = \norm{\vx},
\end{align*}
where in the second equality we changed variables, $\vu=\mR \vv$, where we chose $\mR\in\Real^{k\times k}$ orthogonal so that $\mR^T\frac{\vx}{\norm{\vx}}=(1,0,\ldots,0)^T$, and used the rotation invariance of $\mu$, namely $\mu(\mR \vv) = \mu(\vv)$. In the last equality we used the mean of the folded normal distribution. 
Therefore we get $f(\vx)\approx {\norm{\vx}-r}$.
\end{proof}

\end{document}